\documentclass{article}

\usepackage[final]{neurips_2025}

\usepackage[utf8]{inputenc} 
\usepackage[T1]{fontenc}  
\usepackage{url}         
\usepackage{booktabs}     
\usepackage{amsfonts}      
\usepackage{nicefrac}    
\usepackage{microtype}  
\usepackage{xcolor}     
\usepackage{graphicx}
\usepackage{amsmath}
\usepackage{amssymb}
\usepackage{booktabs}
\usepackage{arydshln} 
\usepackage{adjustbox}
\usepackage{tcolorbox}
\tcbuselibrary{skins}
\usepackage{wrapfig}
\usepackage{multirow}
\usepackage{listings}
\usepackage{subcaption} 
\usepackage{mathtools}
\usepackage{pifont}
\usepackage{cuted}
\usepackage{tabularx}
\definecolor{citecolor}{HTML}{0071bc}
\usepackage[colorlinks, linkcolor=red, colorlinks, anchorcolor=blue, citecolor=citecolor, pagebackref=True]{hyperref}

\usepackage{soul}
\usepackage{diagbox}
\usepackage{makecell}     
\usepackage{array}       
\usepackage{calc}      
\usepackage{nccmath}
\usepackage{xspace}
\usepackage{wrapfig}
\usepackage{paralist}
\usepackage[table]{xcolor}

\usepackage[ruled,vlined]{algorithm2e}

\newcolumntype{Y}{>{\centering\arraybackslash}X} 
\newcolumntype{C}{>{\centering\arraybackslash}X}
\newcommand{\eg}{\textit{e.g.}\xspace}
\usepackage{xcolor}

\newcommand{\tcpgreen}[1]{\textcolor{green!40!black}{\tcp{#1}}}

\title{CQ-DINO: Mitigating Gradient Dilution via Category Queries for Vast Vocabulary Object Detection}

\makeatletter
\usepackage{amsmath} 
\usepackage{amssymb} 
\usepackage{marvosym}

\newcommand{\myfnsymbol}[1]{%
  \expandafter\@myfnsymbol\csname c@#1\endcsname
}
\newcommand{\@myfnsymbol}[1]{%
  \ifcase #1
  \or 1
  \or 2
  \or \TextOrMath{\Letter}{\Letter}
  \or \TextOrMath{\textasteriskcentered}{*}
  \or 3

  \fi
}

\newcommand{\affiliationWHU}{\@myfnsymbol{1}}
\newcommand{\affiliationXHS}{\@myfnsymbol{2}}
\newcommand{\affiliationLJL}{\@myfnsymbol{5}}
\newcommand{\corresponding}{\@myfnsymbol{3}}
\newcommand{\equalcontributor}{\@myfnsymbol{4}}
\makeatother

\author{Zhichao Sun \textsuperscript{\normalfont{\affiliationWHU, \affiliationLJL}} \hspace{2em}\\
\And
Huazhang Hu \textsuperscript{\normalfont{\affiliationXHS}} \hspace{2em}\\
\And
Yidong Ma \textsuperscript{\normalfont{\affiliationXHS}} \hspace{2em}\\
\And
Gang Liu \textsuperscript{\normalfont{\affiliationXHS}} \hspace{2em}\\
\And
Yibo Chen \textsuperscript{\normalfont{\affiliationXHS}} \\
\And
Xu Tang \textsuperscript{\normalfont{\affiliationXHS}}\\
\And
Yao Hu \textsuperscript{\normalfont{\affiliationXHS}} \\
\And
Yongchao Xu \textsuperscript{\normalfont{\affiliationWHU, \affiliationLJL \corresponding}} \\
}

\begin{document}

\renewcommand{\thefootnote}{\myfnsymbol{footnote}}
\maketitle

\footnotetext[3]{Corresponding author: Yongchao Xu <yongchao.xu@whu.edu.cn>}%
\footnotetext[4]{Source Code: \url{https://github.com/FireRedTeam/CQ-DINO}}
\setcounter{footnote}{0}
\renewcommand{\thefootnote}{\arabic{footnote}}
\vspace{-1.5em}
\hspace{2.3em}
\textsuperscript{\normalfont{\affiliationWHU}} \textbf{Institute of Artificial Intelligence, School of Computer Science, Wuhan University}

\hspace{5.3em}
\textsuperscript{\normalfont{\affiliationXHS}} \textbf{Xiaohongshu Inc.}
\hspace{8.3em}
\textsuperscript{\normalfont{\affiliationLJL}} \textbf{Hubei Luojia Laboratory}

\vspace{1em}

\footnotetext[1]{\{zhichaosun, yongchao.xu\}@whu.edu.cn}
\footnotetext[2] {\{huhuazhang, mayidong, tangshen\}@xiaohongshu.com   \{liugang.spl, nemochen89, yaoohu\}@gmail.com}

\begin{abstract}
With the exponential growth of data, traditional object detection methods are increasingly struggling to handle vast vocabulary object detection tasks effectively. We analyze two key limitations of classification-based detectors: \textbf{positive gradient dilution}, where rare positive categories receive insufficient learning signals, and \textbf{hard negative gradient dilution}, where discriminative gradients are overwhelmed by numerous easy negatives. To address these challenges, we propose \textbf{CQ-DINO}, a category query-based object detection framework that reformulates classification as a contrastive task between object queries and learnable category queries. Our method introduces image-guided query selection, which reduces the negative space by adaptively retrieving the top-K relevant categories per image via cross-attention, thereby rebalancing gradient distributions and facilitating implicit hard example mining. Furthermore, CQ-DINO flexibly integrates explicit hierarchical category relationships in structured datasets (\eg, V3Det) or learns implicit category correlations via self-attention in generic datasets (\eg, COCO). Experiments demonstrate that CQ-DINO achieves superior performance on the challenging V3Det benchmark (surpassing previous methods by 2.1\% AP) while maintaining competitiveness on COCO. Our work provides a scalable solution for real-world detection systems requiring wide category coverage.

\end{abstract}
    
\begin{figure}
    \centering
    \includegraphics[width=1\textwidth]{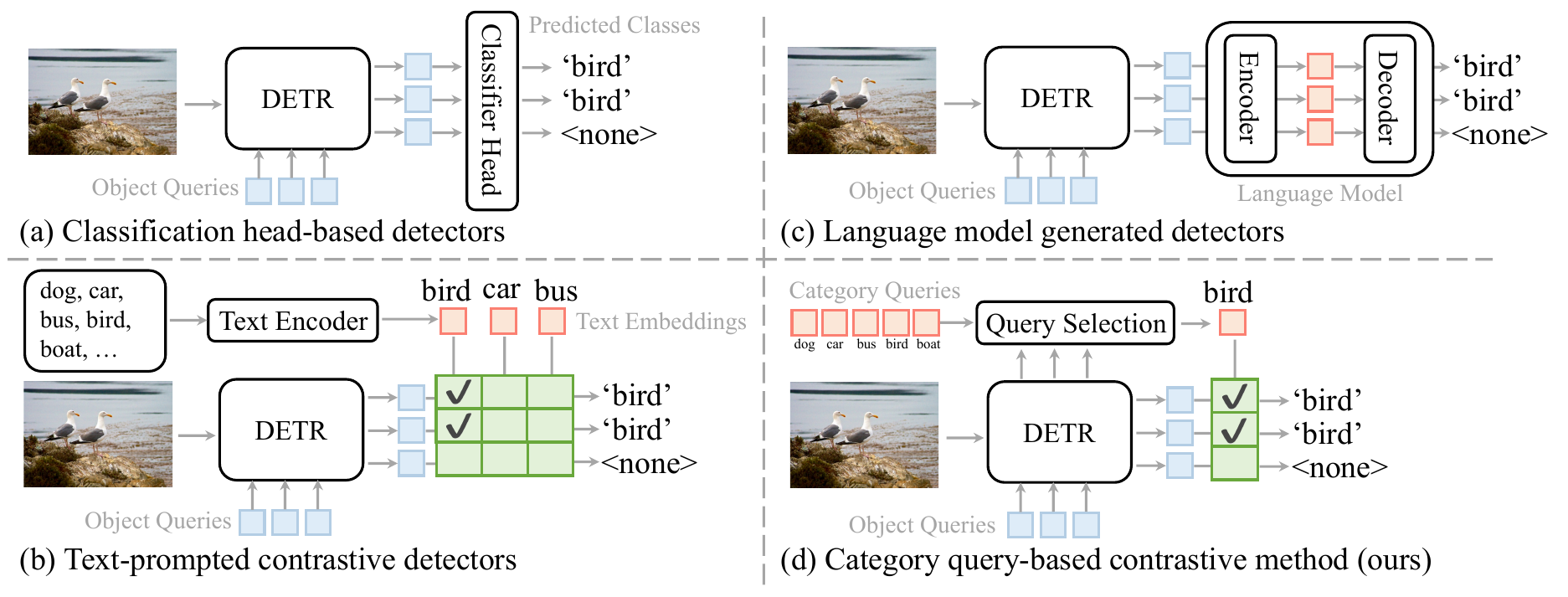}
    \caption{Comparison of category prediction mechanisms for vast vocabulary object detection. (a) Classification head-based detectors with fixed FFN layers face severe optimization challenges with increasing vocabulary size. (b) Text-prompted contrastive detectors leverage VLMs but require multiple inference passes for vast category lists. (c) Language model generated detectors enable open-ended detection but lack control over category granularity. (d) Our proposed CQ-DINO encodes categories as learnable category queries and leverages query selection to identify the most relevant categories in the image, achieving both scalability and improved performance.}
    \vspace{-15pt}
    \label{fig:intro}
\end{figure}
\section{Introduction}
\label{sec:intro}

With the rapid expansion of data, developing a robust AI system capable of large-scale object detection has become essential. This necessity is driven by the increasing complexity and diversity of real-world applications, where AI must manage an extensive vocabulary and dynamic environments. Vast vocabularies inherently present hierarchical category structures, as illustrated by classification datasets like ImageNet~\cite{deng2009imagenet} and Bamboo~\cite{zhang2022bamboo}. Recent detection benchmarks, such as V3Det~\cite{wang2023v3det}, which feature 13,204 object categories organized in hierarchical structures, highlight the magnitude of this challenge. While object detection has witnessed significant advancements~\cite{fasterrcnn,glip,gdino,fu2025llmdet}, scaling effectively to vast vocabularies remains a substantial challenge.
\par
Category prediction mechanisms can be broadly categorized into four types, as illustrated in Fig.~\ref{fig:intro}.
Classification head-based methods employ feed-forward networks (FFNs) with sigmoid activation and Focal Loss~\cite{ross2017focal} optimization. These approaches perform well on benchmarks with limited categories such as COCO~\cite{coco} (80 categories) and Objects365~\cite{shao2019objects365} (365 categories), but face fundamental challenges and scalability issues in vast vocabulary settings.
Text-prompted contrastive methods leverage Vision-Language Models (VLMs) to encode target categories as text inputs, achieving strong open-vocabulary detection~\cite{ovcoco} performance. However, in vast vocabulary scenarios, text input sequences become prohibitively long, necessitating the splitting of category lists across multiple inference passes. This substantially limits their practical scalability and efficiency.
Language model-generated methods approach open-ended~\cite{generateu} object detection by using language models to generate category labels without predefined candidate sets. Although these methods theoretically enable the detection of arbitrary object categories, they generally lack mechanisms to control the granularity of generated labels, which can result in substantial misalignment with practical detection requirements.
\par
In this work, we first systematically analyze the challenges in vast vocabulary detection, focusing particularly on classification-based methods. Our analysis reveals two critical limitations: \textbf{(1) positive gradient dilution}, where the sparse positive categories receive insufficient gradient updates compared to the overwhelming negative categories, and \textbf{(2) hard negative gradient dilution}, where informative hard negative gradients get overwhelmed among numerous easy negative examples.
\par
To tackle these challenges, we introduce \textbf{Category Query-based DINO (CQ-DINO)}, a novel architecture that encodes categories as learnable query embeddings. Our approach centers on image-guided query selection, which identifies relevant categories via category-to-image similarity through cross-attention. The key insight driving our method is that \textbf{\emph{dynamic sparse category selection significantly reduces the negative search space}}. This selection mechanism provides three crucial benefits: (1) balancing the ratio between positive and negative gradients, (2) performing implicit hard mining by selecting the most similar categories, and (3) reducing memory and computational costs, making the framework scalable to extremely large vocabularies. Selected category queries interact with image features to generate object queries. From these object queries, bounding boxes are predicted through a cross-modality decoder. The final classifications are obtained using contrastive alignment between the object queries and the category queries. Unlike traditional classification head-based methods, our category query representation offers greater flexibility by naturally encoding inter-category relationships. For structured datasets with explicit category hierarchies like V3Det~\cite{wang2023v3det}, we leverage the inherent tree structure to construct hierarchical category queries with an adaptive weighting mechanism that balances local and hierarchical features. For datasets without explicit hierarchies (\eg, COCO~\cite{coco}), we employ self-attention mechanisms to implicitly learn the correlations between categories.
\par
We evaluate CQ-DINO on both the vast vocabulary V3Det dataset~\cite{wang2023v3det} and the standard COCO benchmark~\cite{coco}. Our approach surpasses previous state-of-the-art methods on V3Det while maintaining competitive results on COCO compared to DETR-based detectors. Our method benefits from vast vocabulary detection while maintaining competitive results in limited vocabulary scenarios.

\par
Our contributions can be summarized as follows: 
\begin{compactitem}
\item We systematically analyze the challenges in vast vocabulary object detection, identifying positive gradient dilution and hard negative gradient dilution as critical limitations of classification-based methods.
\item We introduce learnable category queries that flexibly encode category correlations with efficient hierarchical tree construction for explicitly modeling category relationships in vast vocabulary scenarios.
\item We develop an image-guided query selection module that dynamically identifies relevant categories per image, effectively addressing the identified limitations while significantly reducing computational complexity.
\end{compactitem}

\section{Related Work}
\label{sec:related}

\subsection{Vast Vocabulary Object Detection}

The progression of object detection benchmarks reflects a steady growth in category vocabulary, evolving from relatively small vocabulary datasets such as PASCAL VOC~\cite{everingham2015pascal} (20 classes) and COCO~\cite{coco} (80 classes), to larger vocabulary benchmarks,  including Objects365~\cite{shao2019objects365} (365 classes) and Open Images~\cite{openimage} (600 classes). Recently, Wang et al.~\cite{wang2023v3det} introduce V3Det, the first vast vocabulary object detection dataset, comprising 13,204 hierarchically structured categories. This unprecedented scale poses significant challenges in terms of scalability and representation.

Recent progress has been driven by the V3Det Challenge~\cite{v3detchallenge}, yielding several methodological advancements. MixPLv2 proposes a semi-supervised framework that combines labeled V3Det data with unlabeled Objects365~\cite{shao2019objects365} images through pseudo-labeling. RichSem-DINO-FocalNet enhances detection robustness by integrating the RichSem-DINO~\cite{RichSem-DINO} framework with a FocalNet-Huge backbone~\cite{FocalNet} pretrained on Objects365. Most recently, Prova~\cite{chen2024comprehensive} introduces multi-modal image-text prototypes specifically optimized for V3Det's fine-grained classification. However, these methods still rely on fundamentally similar classification architectures, raising questions about their effectiveness and scalability for even larger vocabularies beyond tens of thousands of categories.

\subsection{Object Detectors}
\paragraph{Classification Head-based Methods.} Most object detection frameworks employ feedforward networks (FFNs) as fixed classification heads for category prediction.  In two-stage detectors~\cite{fasterrcnn, cascadercnn, he2017mask}, region proposals are generated in a class-agnostic manner by Region Proposal Networks (RPNs), followed by FFN-based region classifiers. By contrast, one-stage detectors~\cite{ross2017focal, fcos, duan2019centernet} predict bounding boxes and categories in a single step. Transformer-based methods, such as DETR~\cite{detr}, reformulate detection as a set prediction problem using learnable queries. Although DETR achieves an elegant end-to-end design, it suffers from slow convergence. Subsequent works~\cite{zhu2021deformable, anchor, dabdetr, dndetr, zhang2023dino, pu2023rank, liu2023detection, hou2024relation} mitigate these limitations. For example, Deformable DETR~\cite{zhu2021deformable} proposes multi-scale deformable attention for sparse spatial sampling. 
DINO~\cite{zhang2023dino} improves performance via contrastive query denoising and mixed query selection.
Despite these architectural advances, current methods fundamentally rely on classification heads with activation functions, typically optimized with Focal Loss~\cite{ross2017focal} or Cross-Entropy Loss~\cite{cross-entropy}. This design inherently constrains scalability and presents optimization challenges when extending to vast vocabulary detection scenarios.

\paragraph{Text-prompted Contrastive Methods.} Vision-language models have advanced the seamless integration of visual and textual modalities for open-vocabulary object detection. These methods encode target categories as text inputs and align visual and textual representations. For instance, GLIP~\cite{glip} pioneers the use of contrastive learning between image regions and textual phrases. Grounding DINO~\cite{gdino} further improves cross-modal alignment through early fusion of vision and textual features. Similarly, DetCLIP~\cite{yao2022detclip} and RegionCLIP~\cite{zhong2022regionclip} leverage image-text pairs with pseudo-labels to enhance region-level semantic understanding and improve generalization. Despite these advances, text-prompted methods face scalability bottlenecks due to the limited capacity of text token inputs during inference. For example, GLIP~\cite{glip} and Grounding DINO~\cite{gdino} restrict input prompts to approximately 128 tokens per pass, allowing for about 40 categories simultaneously. Thus, detecting all categories in a vast vocabulary benchmark like V3Det~\cite{wang2023v3det} (13,204 classes) would require over \emph{331 sequential inference passes per image.} This constraint renders current text-prompted methods computationally inefficient and impractical for real-time or large-scale detection.

\paragraph{Language Model-Generated Methods.} Advances in multimodal large language models (MLLMs) have inspired detection methods leveraging their visual understanding and generative abilities. Some MLLMs~\cite{internvl, bai2023qwen, gemini} show preliminary object detection abilities but exhibit limited localization precision and recognition granularity. To mitigate these limitations, recent works~\cite{generateu, fu2025llmdet, jiang2024chatrex, wang2023v3det, glee} use LLMs primarily as category generators rather than direct detectors. For instance, GenerateU~\cite{generateu} employs a T5-based decoder~\cite{t5} to generate category names from visual features, reframing detection as text generation. Similarly, LLMDet~\cite{fu2025llmdet} and ChatRex~\cite{jiang2024chatrex} utilize instruction-tuned LLMs to predict object categories from image features. While these generative approaches enable open-ended detection, they often produce inconsistencies due to limited controllability over label granularity. For example, given an image region of a ``Persian cat'', the model may generate the generic term ``cat'', causing semantic ambiguity and reduced accuracy for fine-grained detection tasks.

\subsection{Category Query-based Methods}
Learnable queries was popularized in computer vision by DETR~\cite{detr}, marking a paradigm shift from fixed architectural components to task-adaptive representations. Queries serve as learnable embeddings that interact with the visual feature space, enabling the model to capture complex, task-specific patterns. This design has since been adopted in diverse domains, including classification~\cite{liu2021query2label}, segmentation~\cite{cheng2021per}, and multimodal learning~\cite{li2023blip}. Among these developments, \textit{category queries} represent an innovation introduced by Query2Label~\cite{liu2021query2label}. Rather than relying on fixed classification heads, Query2Label proposed learnable category embeddings to capture category-specific features. Subsequent works such as ML-Decoder~\cite{ridnik2023ml} have outperformed conventional classification methods. The effectiveness of category queries in classification motivated their extension to dense prediction tasks. For example, CQL~\cite{xie2023category} applies them to human–object interaction classification, ControlCap~\cite{zhao2024controlcap} uses them for semantic guidance in region captioning, and RankSeg~\cite{he2022rankseg} integrates them into semantic segmentation, dynamically selecting the top-$k$ most relevant classes during inference. This selective querying reduces the effective search space, improving computational efficiency and segmentation accuracy. While prior works have explored category queries in various contexts, our work addresses a fundamentally different challenge specific to vast vocabulary scenarios. We provide, to the best of our knowledge, the first systematic theoretical analysis of \emph{gradient dilution} issues that arise when dealing with vast category vocabularies, which motivates our image-guided query selection design. Moreover, we introduce a hierarchical tree construction strategy that explicitly models category correlations, enabling effective reasoning over deep semantic hierarchies in vast vocabulary datasets.

\begin{wrapfigure}{r}{0.5\textwidth}
    \centering
    \includegraphics[width=1\linewidth]{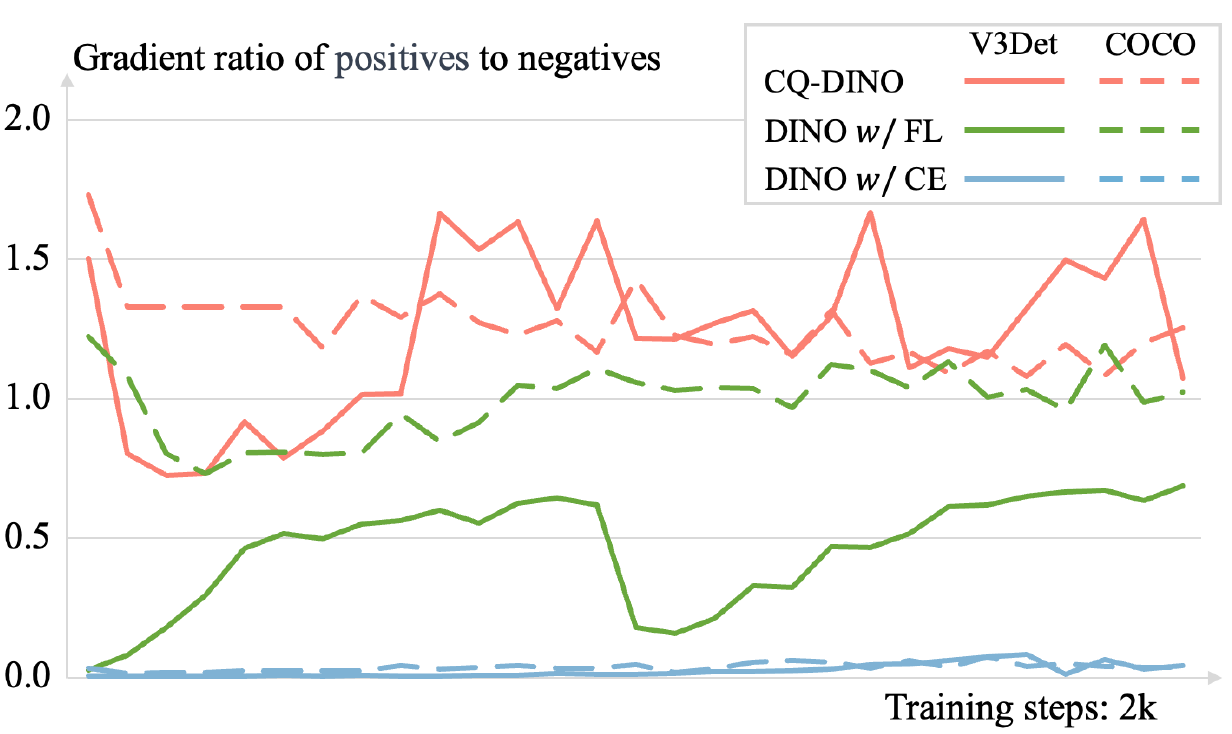}
\vspace{-20pt}
\caption{\small{Positive-to-negative gradient ratio comparing CQ-DINO against DINO with Focal Loss (FL) and Cross-Entropy Loss (CE) on V3Det and COCO datasets, showing the initial 2k training iterations where differences are most evident.}}
    \label{fig:ratio}
\end{wrapfigure}

\section{Method}
\label{method}
\subsection{Challenges in Vast Vocabulary Detection}
\label{sec:challenge}
Existing methods for vast vocabulary object detection with $C$ categories ($C > 10^4$), particularly those employing sigmoid-based classifiers with Focal Loss~\cite{ross2017focal}, face fundamental optimization challenges. We systematically analyze these issues through a simplified formulation using the Cross-Entropy Loss with sigmoid activation, revealing two critical limitations:
\par
\paragraph{1) Positive Gradient Dilution.} In vast vocabulary detection, the gradient signal for positive classes is overwhelmed by the aggregated negative gradients. Let $z_c$ denote the logit for class $c$ and $y_c \in \{0, 1\}$ be its ground-truth label. The gradient of the Cross-Entropy Loss $\mathcal{L}$ with respect to $z_c$ is:
\begin{equation}
    \label{eq:gradient} 
    \nabla_{z_c}\mathcal{L} = \sigma (z_c)-y_c~,
\end{equation}
 where $\sigma(\cdot)$ is the sigmoid function. For a positive class $c^+$ ($y_{c^+}=1$), the gradient magnitude is $\lvert\nabla_{z_{c^+}}\mathcal{L}\rvert = 1-\sigma(z_{c^+})$, while for negatives $c^-$ ($y_{c^-}=0$), it is $\lvert\nabla_{z_{c^-}}\mathcal{L}\rvert = \sigma(z_{c^-})$.
 \par
 The total negative gradient magnitude grows linearly with the category count $C$:
\begin{equation}
    \label{softmax}
    ||\nabla_{z_{c^+}}\mathcal{L}|| \ll \sum_{c^-\neq c^+}^C ||\nabla_{z_{c^-}}\mathcal{L}|| \approx (C-1)\cdot \epsilon  ,
\end{equation}
where $\epsilon = \mathbb{E}[\sigma(z_{c^-})]$ represents the average activation probability of negative classes. The positive-to-negative gradient ratio $\rho$ becomes:
 \begin{equation}
\label{eq:gradient_ratio} 
    \rho = \frac{||\nabla_{z_{c^+}}\mathcal{L}||}{\sum_{c^-\neq c^+}^C ||\nabla_{z_{c^-}}\mathcal{L}||} \approx \frac{1-\sigma(z_{c^+})}{(C-1)\cdot \mathbb{E}[\sigma(z_{c^-})]}  \propto \frac{1}{C\cdot \epsilon}.
\end{equation}
During early training stages, $\epsilon$ retains a non-negligible value. Since $C~\text{exceeds}~10^4$ in vast category scenarios, $\rho \to 0$, causing positive gradients to be suppressed relative to the cumulative negative gradients. This fundamentally hinders the model's ability to learn from positive examples.

\paragraph{2) Hard Negative Gradient Dilution.} The massive negative space leads to gradient dominance by easily classified negatives rather than informative hard negatives. Let $\mathcal{H}$ denote the set of hard negative classes with $\mathbb{E}[\sigma(z_{c^h})] = \epsilon^h$ for $c^h \in \mathcal{H}$. The ratio of hard negative gradients to total negative gradients is:
\begin{equation} 
\label{eq:hard_ratio} 
\eta = \frac{\sum_{c^h \in \mathcal{H}}|\nabla_{z_{c^h}}\mathcal{L}|}{\sum_{c^- \neq c^+}^{C} |\nabla_{z_{c^-}}\mathcal{L}|} \approx \frac{N_h}{C} \cdot \frac{\epsilon^h}{\epsilon}, \end{equation} 
where $N_h$ is the number of hard negatives. As $C~\text{exceeds}~10^4$, $\eta \to 0$ due to the $\frac{1}{C}$ term, making hard negatives diluted in gradient updates.

\par
Fig.~\ref{fig:ratio} demonstrates these theoretical challenges. The gradient ratio for the V3Det dataset (13,204 classes) is lower than for the COCO dataset (80 classes), revealing the inherent difficulty in vast vocabulary object detection. While Focal Loss partially mitigates these issues by down-weighting easy negatives, the gradient ratio for V3Det remains around 0.5 compared to approximately 1.0 for COCO, indicating that gradient imbalance persists despite these improvements. We provide a comprehensive experimental analysis of Focal Loss performance and limitations in Sec.~\ref{sec:discussion}.

\begin{figure}[t]
    \centering
    \includegraphics[width=1.0\linewidth]{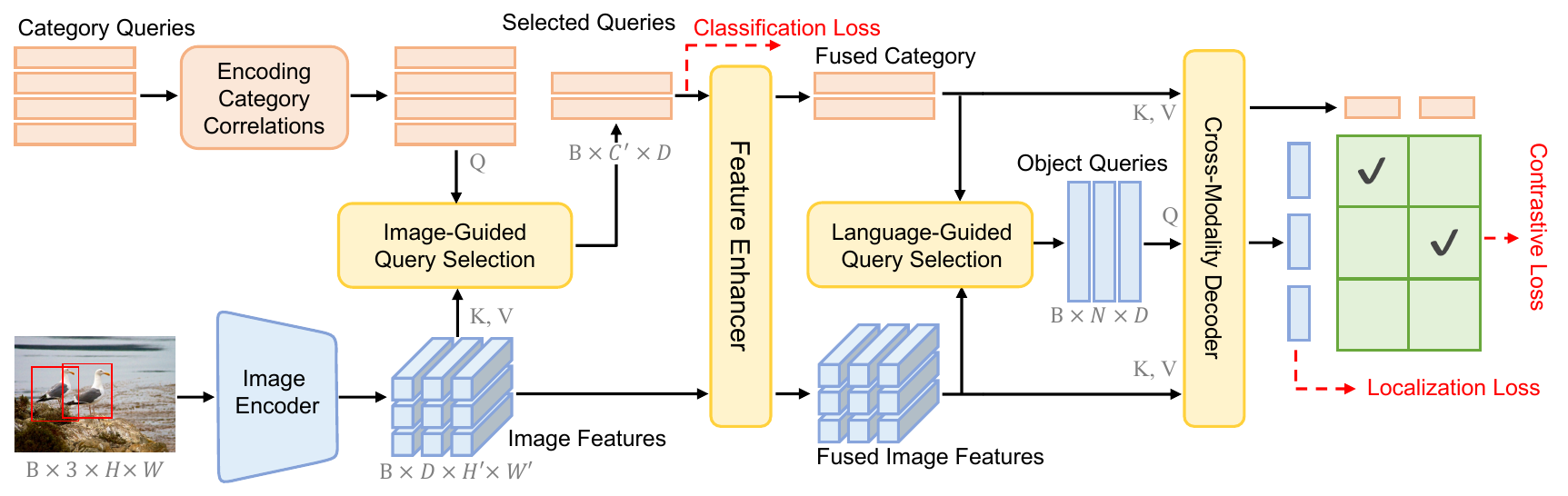}
 
\caption{Overview of the CQ-DINO framework for vast vocabulary object detection. Key components: (1) Learnable category queries enhanced with hierarchical tree construction for semantic relationship modeling; (2) Image-guided query selection that identifies the most relevant category queries; (3) Feature enhancer and cross-modality decoder (adapted from GroundingDINO~\cite{gdino}), processing object queries with contrastive alignment between object and selected category queries.}
  
    \label{fig:main}
\end{figure}

\subsection{CQ-DINO}
\label{overall}
Our key insight is that dynamically selecting a sparse category subset $S \subset \{1, \dots, C\}$ simultaneously addresses both gradient dilution challenges through gradient magnitude rebalancing and adaptive hard negative mining. As shown in Fig.~\ref{fig:main}, CQ-DINO consists of three key components:
\\
1) Learnable category queries with correlation encoding. We initialize learnable category queries $Q_{cat} \in  \mathbb{R}^{B\times C\times D}$ using the OpenCLIP~\cite{clip} text encoder, where $B$ is the batch size, $C$ denotes the total number of categories, and $D$ is the embedding dimension. These category queries enable flexible encoding of category correlations through self-attention or hierarchical tree construction (Sec.~\ref{sec:tree_struct}).\\
2) Image-guided query selection. Given image features $F_{img} \in \mathbb{R}^{B\times D\times H' \times W'}$ from the image encoder, we compute similarities between $F_{img}$ and $Q_{cat}$ through multi-head cross-attention modules. 
For each image, we select the top-$C'$ most relevant queries ($C' \ll C$, typically $C'=100$ for $C>10^4$), ensuring that the target class $c^+$ and its most confusing negative classes are preserved. This selection process rebalances the gradients and implicitly performs hard negative mining.\\
3) Feature enhancer and cross-modality decoder. The selected category queries $Q_{cat}' \in \mathbb{R}^{B\times C'\times D}$ and image features $F_{img}$ are processed through GroundingDINO~\cite{gdino} components. First, the feature enhancer module fuses the category queries and image features. Next, object queries are generated through language-guided query selection. Finally, detection outputs are produced by the cross-modality decoder and contrastive alignment between object queries and selected category queries.

\begin{figure}[t]
\begin{minipage}[t]{0.49\textwidth}
    \centering
    \includegraphics[width=1\linewidth]{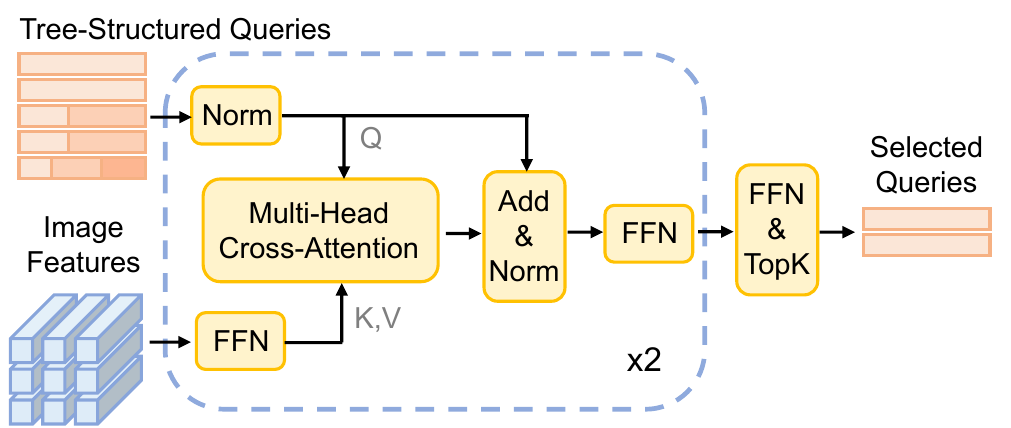}
\caption{Illustration of image-guided query \\ selection module.}
    \label{fig:igqs}
\end{minipage}
\hfill
\begin{minipage}[t]{0.49\textwidth}
    \centering
    \includegraphics[width=1\linewidth]{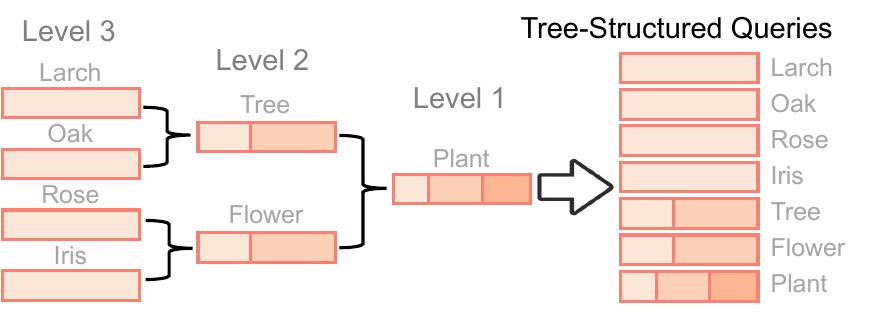}
\caption{Hierarchical tree construction for \\ category queries.}
    \label{fig:tree}
\end{minipage}
\end{figure}

\subsection{Image-Guided Query Selection}
\label{query_section}
The core innovation of CQ-DINO is our image-guided category selection module, illustrated in Fig.~\ref{fig:igqs}. This module employs cross-attention between category queries $Q_{cat}$ and image features $F_{img}$. Here, $Q_{cat}$ serves as queries (Q), while $F_{img}$ provides keys (K) and values (V). The cross-attention layer establishes category-to-image correlations through similarity computation. Then, we apply TopK selection to retain only the top-$C'$ categories ($C'\ll C$) based on activation values. We supervise this selection using Asymmetric Loss~\cite{ridnik2021asymmetric}, which serves as a multi-class classification loss.

\par
The selection rebalances the positive-to-negative gradient ratio. Let $\rho$ and $\rho'$ denote the original and revised positive-to-negative gradient ratios, respectively:
\begin{equation}
\frac{\rho'}{\rho} = \frac{\sum_{c^-}^C ||\nabla_{z_{c^-}}\mathcal{L}||}{\sum_{c^-\in S}^{C'} ||\nabla_{z_{c^-}}\mathcal{L}||} \approx \frac{C}{C'}.
\end{equation}
For a typical setting ($C>10^4,~C'=100$), this achieves a $100\times$ gradient rebalancing factor. Our design provides three benefits:
\textbf{1) Gradient rebalancing.} By filtering out easy negative categories, our selection module improves the influence of gradients from positive examples.
\textbf{2) Adaptive hard negative mining.} The selection mechanism ensures that retained negative categories exhibit high semantic relevance to the image content, naturally implementing hard negative mining.
\textbf{3) Scalable computation.} Processing only $C'$ categories reduces memory consumption and computational cost, making the framework scalable to extremely large vocabularies.

\subsection{Encoding Category Correlations}
\label{sec:tree_struct}
A key advantage of our category query approach is the capacity to model complex semantic relationships among categories, which is difficult for traditional classification head-based methods to achieve. We propose two complementary strategies for encoding category correlations.
\par
\noindent\textbf{Explicit Hierarchical Tree Construction.} \quad For hierarchical datasets like V3Det~\cite{wang2023v3det}, we introduce hierarchical tree construction, as shown in Fig.~\ref{fig:tree}. The process begins at leaf nodes and progressively integrates hierarchical information upward through the category tree. For each parent node $v$ with children $\mathcal{C}(v)$, its new query $Q_v^{\prime}$ is a combination of its original query $Q_v$ and the mean pooling of all direct child nodes. Leaf nodes retain their original features directly, since they have no children. 
\begin{equation} 
Q_v^{\prime} = (1-\alpha_v) \cdot Q_v + \alpha_v \cdot \frac{1}{|\mathcal{C}(v)|} \sum_{c \in \mathcal{C}(v)} Q_c,
\label{eq:tree_aggregation_v} 
\end{equation}
where $\alpha_v \in [0,1]$ balances local and hierarchical features.  
\begin{equation}
    \alpha_v = w\left(1 + \frac{\log(n_v + 1)}{\log(N_{\text{max}} + 1)}\right), 
\end{equation}
where $n_v$ is the child count for node $v$, $N_\text{max}$ is the maximum child count across the tree, and $w \in [0, 0.5]$ is a hyperparameter (default: 0.3). This adaptive weight $\alpha_v$ ensures that parent nodes with more descendants incorporate more collective knowledge, while nodes with fewer children maintain stronger individual semantics.

\par
Building upon this structural design, we introduce a masking strategy during the classification loss computation to mitigate hierarchical ambiguity. If any child category exists in the ground truth, its parent nodes are excluded from the classification loss. This prevents conflicting supervision signals for semantically related categories, such as suppressing ``vehicle'' when ``car'' is annotated.

\par
\noindent\textbf{Implicit Relation Learning.} \quad For categories without an explicit hierarchical structure, we employ a self-attention mechanism to learn category relationships. This allows semantically related categories to influence each other's representations based on learned attention patterns.

\section{Experiments}

\subsection{Datasets and Implementation Details}
We conduct experiments on two detection benchmarks:
(1) V3Det~\cite{wang2023v3det}: a vast vocabulary detection dataset containing 13,204 categories, with 183k training and 30k validation images. This is our primary benchmark for evaluating vast vocabulary detection.
(2) COCO \texttt{val2017}~\cite{coco}: a standard benchmark dataset with 80 object categories, comprising 118k training and 5k validation images. We include this dataset to verify the effectiveness of our method in limited vocabulary scenarios.

\begin{table}[t]
  \centering
  \small
\caption{Comparison with state-of-the-art methods on the V3Det validation set. Best results in each group are highlighted in \textbf{bold}.}
\label{tab:v3det}
  \begin{tabular}{@{} l @{} c c c c c @{}}
    \toprule
    Method & Epochs & Backbone & $AP$ & $AP_{50}$ & $AP_{75}$  \\ 

\midrule

ATSS ~\cite{atss}  & 24 & Swin-B & 7.6 & 8.9 & 8.0 \\
FCOS~\cite{fcos}  & 24 & Swin-B & 21.0 & 24.8 & 22.3 \\
Faster R-CNN~\cite{fasterrcnn}  & 24 & Swin-B & 37.6 & 46.0 & 41.1 \\
CenterNet2~\cite{centernet2} & 24 & Swin-B & 39.8 & 46.1 & 42.4 \\
Cascade R-CNN~\cite{cascadercnn}  & 24 & Swin-B & 42.5 & 49.1 & 44.9 \\
Deformable DETR~\cite{zhu2021deformable} & 50 & Swin-B & 42.5 & 48.3 & 44.7 \\

DINO~\cite{zhang2023dino} & 24 & Swin-B & 42.0 & 46.8 & 43.9 \\
Prova~\cite{chen2024comprehensive} & 24 & Swin-B & 44.5 & 49.9 & 46.6 \\  
 CQ-DINO (Ours) & 24 & Swin-B  & \textbf{46.3} & \textbf{51.5} & \textbf{48.4} \\

\midrule
DINO~\cite{zhang2023dino} & 24 & Swin-B-22k & 43.4 & 48.4 & 45.4 \\
Prova~\cite{chen2024comprehensive} & 24 & Swin-B-22k & 50.3 & 56.1 & 52.6 \\

CQ-DINO (Ours) & 24 & Swin-B-22k  &\textbf{ 52.3 }&\textbf{ 57.7} &\textbf{ 54.6} \\

\midrule
DINO~\cite{zhang2023dino} & 24 & Swin-L & 48.5 & 54.3 & 50.7 \\
Prova~\cite{chen2024comprehensive} & 24 & Swin-L & 50.9  & 57.2  & 53.2  \\
CQ-DINO (Ours) & 24 & Swin-L  & \textbf{53.0} & \textbf{58.4} & \textbf{55.4} \\

\bottomrule

\end{tabular}

\end{table}

\begin{table}[t]

\begin{minipage}[t]{0.53\textwidth}
  \centering
\caption{Comparison between the proposed CQ-DINO and state-of-the-art DETR variants on COCO~\texttt{val2017}, reporting the best results as provided by their respective original papers.}
\label{tab:coco}
\begin{adjustbox}{width=\linewidth,valign=t}
  \begin{tabular}{@{}l @{} c @{\hspace{2pt}} c @{\hspace{2pt}} c@{\hspace{2pt}} c@{\hspace{2pt}} c @{\hspace{2pt}} c@{}}
    \toprule
    Method & Epochs & Backbone & $AP$ & $AP_S$ & $AP_M$ & $AP_L$  \\ 

\midrule
H-Def-DETR~\cite{hfdetr} & 36  & Swin-L  & 57.1  & 39.7  & 61.4 & 73.4  \\
Relation-DETR~\cite{hou2024relation}   & 12 & Swin-L & 57.8 & 41.2 & 62.1 & \textbf{74.4} \\
DINO~\cite{zhang2023dino} & 36 & Swin-L  & 58.0 & 41.3 & 62.1 & 73.6 \\
Rank-DETR~\cite{pu2023rank} & 36 & Swin-L  & 58.2 & 42.4 & \textbf{62.2} & 73.6 \\
CQ-DINO (ours) & 24 & Swin-L  & \textbf{58.5} & \textbf{42.5} & 62.1 & 74.0 \\
\bottomrule

\end{tabular}

\end{adjustbox}
\end{minipage}
\hfill
\begin{minipage}[t]{0.44\textwidth}
\caption{Performance of Open-world methods on V3Det dataset. We report zero-shot performance using their strongest models. * * indicates finetuned results.}
\label{tab:generate}
\begin{adjustbox}{width=\linewidth,valign=t}
  \begin{tabular}{@{} l @{} c @{\hspace{2pt}} c@{\hspace{2pt}} c @{\hspace{2pt}}c @{}}
    \toprule
    Method  & Backbone & $AP$ & $AP_{50}$ & $AP_{75}$  \\ 

\midrule
GenerateU~\cite{generateu} & Swin-L\&T5-B & 0.4 & 0.5 & 0.4 \\
ChatRex~\cite{jiang2024chatrex} & Swin-L\&LLM-7B & 1.3 & 1.5 & 1.4 \\
*GenerateU*~\cite{generateu} & Swin-L\&T5-B & 21.8 & 27.2 & 22.1 \\
\midrule
DINO~\cite{zhang2023dino}  & Swin-L & 48.5 & 54.4 & 50.7 \\
CQ-DINO (Ours)  & Swin-L  & \textbf{53.0} & \textbf{58.4} & \textbf{55.4} \\
\bottomrule
\end{tabular}
\end{adjustbox}
\end{minipage}

\end{table}

 Experiments are conducted on 8 A100-40G GPUs with a total batch size of 16, unless otherwise specified. Baseline configurations are used for fair comparison. Three Swin Transformer~\cite{liu2021Swin} variants serve as backbones: Swin-B (ImageNet-1k~\cite{deng2009imagenet} pre-trained), Swin-B-22k (ImageNet-22k pre-trained), and Swin-L (ImageNet-22k pre-trained). Category queries are initialized using CLIP-ViT-L~\cite{clip} text embeddings. To match dataset vocabularies, we employ 100 category queries for V3Det and 30 for COCO, aligned with their category sizes. The hierarchical tree structure is derived from V3Det's category taxonomy. For implicit relation learning, we use an 8-head self-attention module.
 
\par
The training objective combines multiple loss terms with the following weights: classification loss (Asymmetric Loss~\cite{ridnik2021asymmetric}, weight=1.0), contrastive alignment (Focal Loss~\cite{ross2017focal}, weight=1.0), bounding box regression (L1 Loss, weight=5.0), and GIoU Loss~\cite{giou} (weight=2.0), as in GroundingDINO~\cite{gdino}. Hungarian matching is used, following GroundingDINO, with identical matching costs for object-to-query assignment. To stabilize training, we use a two-stage approach: first, pre-training category queries, image encoder, and image-guided query selection for 10 epochs to establish high initial target category recall; then, fine-tuning the full detection pipeline in the second stage.

\subsection{Experimental Results}

\noindent\textbf{Performance on Vast Vocabulary Detection}.\quad Tab.~\ref{tab:v3det} presents our comparison with state-of-the-art methods on the V3Det~\cite{wang2023v3det} benchmark. CQ-DINO consistently outperforms all previous approaches across different backbone configurations. With the Swin-B backbone, CQ-DINO achieves 46.3\% AP, outperforming general detection methods like Deformable DETR \cite{zhu2021deformable} by 3.8\% AP and DINO \cite{zhang2023dino} by 4.3\% AP. More importantly, CQ-DINO surpasses Prova~\cite{chen2024comprehensive}, a specialized vast vocabulary detection method, by 1.8\% AP. When integrated with the Swin-B-22k backbone, CQ-DINO achieves 52.3\% AP, outperforming Prova by 2.0\% AP. With the Swin-L backbone, CQ-DINO achieves 53.0\% AP. The consistent improvements across different backbones demonstrate that CQ-DINO effectively addresses vast vocabulary detection challenges.

\noindent\paragraph{Performance on Standard Detection Benchmark}. Tab.~\ref{tab:coco} compares our method with state-of-the-art approaches on COCO \texttt{val2017}. Despite being primarily designed for vast vocabulary scenarios, CQ-DINO achieves competitive performance, reaching 58.5\% AP, which is comparable to recent DETR-based methods. We report the best results from their original papers to ensure a fair comparison. The competitive performance of CQ-DINO is mainly due to the proposed gradient rebalancing and adaptive hard mining strategies.

\begin{table}[t]
\caption{Ablation study on the effectiveness of encoding category correlations and image-guided query selection components in CQ-DINO on V3Det dataset with Swin-Base-22k backbone. ``--'' denotes unavailable $AR^C$ metrics due to absence of query selection.}
\label{tab:component_ablation}
\small
  \centering
  \begin{tabular}{ c | c |lll }
    \toprule
Encoding Category Correlations & Image-guided Query Selection & $AP$ & $AR^C$  & FPS \\
    \midrule 
    \phantom{\checkmark} & \phantom{\checkmark}  & 47.3 & -- & 0.7 \\
    Hierarchical Tree construction & \phantom{\checkmark}  &  49.4~\tiny{($\uparrow2.1$}) & -- & 0.6~\tiny{($\downarrow0.1$})  \\
 \midrule

     & \checkmark & 51.1  & 80.9 & \textbf{10.8} \\
   Self-Attention module & \checkmark & 51.3~\tiny{($\uparrow0.2$}) & 75.5~\tiny{($\downarrow5.4$}) & 10.4~\tiny{($\downarrow0.4$}) \\

   Hierarchical Tree Construction  & \checkmark & \textbf{52.3}~\tiny{($\uparrow1.2$}) & 83.3~\tiny{($\uparrow2.4$}) & 10.6~\tiny{($\downarrow0.2$}) \\

  \bottomrule

\end{tabular}

\end{table}

\subsection{Ablation Studies}
We conduct ablation studies to evaluate the effectiveness of each component in CQ-DINO. Unless otherwise stated, all experiments are performed on the V3Det dataset using a Swin-B-22k backbone.

\noindent\paragraph{Effect of Each Component in CQ-DINO.}  Table~\ref{tab:component_ablation} presents the contribution of each component in terms of average precision ($AP$) and category-level average recall with selected queries ($AR^C$). To establish a baseline without image-guided query selection, we conduct experiments on 8 H800-80G GPUs due to memory constraints. Notably, incorporating image-guided query selection increases FPS from 0.7 to 10.8, highlighting its effectiveness in alleviating memory bottlenecks and substantially enhancing inference efficiency. Furthermore, this component addresses the issue of gradient dilution and leads to a substantial AP improvement, from 47.3\% to 51.1\% in detection performance.\par

Explicit hierarchical modeling via tree construction leads to an improvement of 1.2\% AP and 2.4\% $AR^C$. In contrast, employing self-attention  for implicit relationship modeling achieves a marginal increase of 0.2\% AP but reduces $AR^C$. This is due to difficulties in learning complex relationships across 13k+ categories.  Notably, the tree construction method introduces zero additional parameters with only a 0.2 FPS overhead, while the self-attention approach adds 2.36M extra parameters and a 0.4 FPS reduction in inference speed.
\par

While Tab.~\ref{tab:component_ablation} shows that explicit tree construction outperforms self-attention on V3Det's vast category space, we conduct further experiments on COCO (Table~\ref{tab:sa}). The results reveal that self-attention contributes meaningful 0.2\% AP and 0.9\% $AR^C$ improvements on datasets with fewer categories. Both experiments validate the effectiveness of encoding category correlations, with the optimal approach depending on the scale of the category space.

\begin{table}[t]
\begin{minipage}[t]{0.51\textwidth}
  \centering
\caption{Ablation study on self-attention module (SA) in CQ-DINO on the COCO dataset.}
\label{tab:sa}
\begin{adjustbox}{width=\linewidth,valign=t}

  \begin{tabular}{@{}l@{\hspace{1pt}}| @{\hspace{1pt}}l@{\hspace{3pt}} l@{\hspace{3pt}} l @{}}
    \toprule
    Method    & $AP$ & $AR^C$   &  Params (M) \\
    \midrule 
CQ-DINO \textit{w/o} SA  & 58.3  & 98.2 &  \textbf{244.3}\\
CQ-DINO \textit{w/} SA & \textbf{58.5} \tiny{($\uparrow 0.2$)} & \textbf{99.1} \tiny{($\uparrow 0.9$)} &  246.7 \tiny{($+ 2.4$)}\\
  \bottomrule
\end{tabular}
\end{adjustbox}
\end{minipage}
\hfill
\begin{minipage}[t]{0.46\textwidth}
  \centering
\caption{Ablation study on adaptive weighting in tree construction.}
\label{tab:weight}
\begin{adjustbox}{width=\linewidth,valign=t}
\begin{tabular}{@{}l@{\hspace{4pt}}| c@{\hspace{6pt}} c@{\hspace{6pt}} c @{\hspace{6pt}}c@{}}
    \toprule
    Method & $AP$ & $AP_{50}$ & $AP_{75}$  &  $AR^{C}$ \\
    \midrule 
Fixed weight (0.5)  & 51.9 & 57.4 & 54.3   & 82.3 \\
Ours ($\alpha_v$) & \textbf{52.3}  & \textbf{57.7} & \textbf{55.4}   & \textbf{83.3} \\
  \bottomrule
\end{tabular}
\end{adjustbox}
\end{minipage}
\end{table}

\paragraph{Effectiveness of Adaptive Weighting in Tree Construction.} 
Tab.~\ref{tab:weight} shows the importance of our adaptive weighting strategy compared to a fixed weight of 0.5. This approach adjusts weights based on the varying number of child categories for each parent node in the hierarchy. Results show that our adaptive approach outperforms fixed weighting.\par

\subsection{Discussion}
\label{sec:discussion}
\noindent\textbf{Scalability of CQ-DINO.}\quad
Tab.~\ref{tab:scalability} compares the scaling efficiency of CQ-DINO and DINO for vast vocabulary detection. CQ-DINO requires only 0.8K parameters per category, representing a 62\% reduction from DINO's 2.1K parameters. For runtime memory consumption, CQ-DINO uses 2.7KB CUDA memory per category. To evaluate practical scalability limits, we test the maximum category support on an A100-40G GPU using the Swin-B-22k backbone with a single $800\times1333$ resolution input image. CQ-DINO supports detection of up to 130k categories, surpassing DINO's 100k limit. These experiments validate that CQ-DINO enables applications with extremely large vocabularies. \par

\begin{table}[t]
\begin{minipage}[t]{0.49\textwidth}
  \centering
\caption{Scalability comparison of CQ-DINO with DINO on A100 40G GPU using Swin-B-22k backbone, showing per-category parameters (K), CUDA memory consumption (kB), and maximum supported category capacity (k).}
\label{tab:scalability}
\begin{adjustbox}{width=\linewidth,valign=t}
\begin{tabular}{ l@{\hspace{3pt}} |c@{\hspace{3pt}} c@{\hspace{3pt}} c@{\hspace{3pt}} @{}}
\toprule
\multirow{2}{*}{Method} & Params/Cat. & Memory/Cat. & Max Cats. \\

 & (K) & (kB) & (k) \\ 
 
 \midrule
DINO~\cite{zhang2023dino} & 2.1 & 8.9 & 100 \\
CQ-DINO & 0.8 & 2.7 & 130 \\
\bottomrule
\end{tabular}
\end{adjustbox}

\end{minipage}
\hfill
\begin{minipage}[t]{0.49\textwidth}
\caption{Focal Loss parameter analysis in DINO using Swin-B-22k backbone. AP scores (\%) compare different $\alpha$ and $\gamma$ combinations. Dashes ``--'' indicate unstable training configurations.}
\vspace{2pt}
\label{tab:focal}
\begin{adjustbox}{width=1\linewidth,valign=t}
\setlength{\tabcolsep}{12pt} 

\begin{tabular}{c|cccc@{}}
\toprule
\diagbox{$\gamma$}{$\alpha$} & 0.25 & 0.35 & 0.50 & 0.75 \\
\midrule
2 & 43.4 & 45.1 & \textbf{47.4}  & -- \\
3 & 43.7 & 43.9 & 45.1 & --  \\
5 & -- & -- & --  &  -- \\
\bottomrule
\end{tabular}
\end{adjustbox}
\end{minipage}
\end{table}

\paragraph{Limitations of Generation-based Methods.}
We evaluate generation-based methods on the V3Det dataset in Tab.~\ref{tab:generate}. Following the evaluation protocol from GenerateU~\cite{generateu}, we compute semantic similarity between generated category embeddings and V3Det category embeddings, selecting the highest similarity match as the final prediction. Our experiments reveal poor performance: GenerateU~\cite{generateu} achieves only 0.4\% AP, while the more recent ChatRex~\cite{jiang2024chatrex} achieves just 1.3\% AP. This highlights a fundamental limitation: \textbf{\emph{generation-based methods struggle to control the granularity of generated categories, creating semantic misalignments with the specific requirements of detection tasks.}} Furthermore, even when we finetune GenerateU on V3Det data, the resulting performance (21.8\% AP) still exhibits a substantial gap compared to classification-head methods.\par

\paragraph{Focal Loss Parameters Analysis under Gradient Dilution.}
As discussed in Sec.~\ref{sec:challenge}, vast vocabulary detection suffers from gradient dilution challenges. Focal Loss (FL)~\cite{ross2017focal} mitigates this via adaptive weighting with hyperparameters $\alpha$ and $\gamma$. Theoretically, $\alpha$ balances positive/negative sample contributions, and increasing $\alpha$ enhances the model's ability to handle more categories. The factor $\gamma$ focuses learning on hard negatives, where increasing $\gamma$ improves the mining of hard examples. Tab.~\ref{tab:focal} evaluates FL configurations on DINO. The default setting~\cite{wang2023v3det} ($\alpha=0.25$, $\gamma=2$) achieves 43.4\% AP. Tuning these hyperparameters reveals that training is unstable when $\gamma \geq 5 $ or $\alpha \geq 0.75 $. Notably, the optimal configuration ($\gamma = 2 $ and $\alpha = 0.5 $) achieves 47.4\% AP, surpassing the baseline by 4.0\%. Nonetheless, this is still lower than the 52.3\% AP achieved by our CQ-DINO, suggesting that FL hyperparameter tuning, while beneficial, leaves room for further improvement. Interestingly, when examining parameter transferability across architectures in Appendix Tab.~\ref{tab:dino}, we find that this optimal setting does not generalize well. Applying the  Swin-B-22k optimal parameters  ($\alpha=0.5$, $\gamma=2$) to Swin-B degrades performance by 3.3 \% AP relative to its default setting.  However, they increase the performance by 4.0\% AP and 1.6\% AP for Swin-B-22k and Swin-L backbones, respectively. These findings suggest that while Focal Loss is effective in addressing the gradient dilution challenge, optimal hyperparameter selection remains architecture-dependent and requires careful tuning.

\subsection{Limitation}

Although CQ-DINO improves vast vocabulary object detection, several limitations remain. First, detection performance is influenced by the recall of the category query selection. Fortunately, CQ-DINO achieves 83.3\% $AR^C$. Appendix Tab.~\ref{tab:number} shows that increasing the number of category queries improves $AR^C$ but does not lead to higher $AP$. This indicates, in most cases, 83.3\% $AR^C$ is not a primary bottleneck. Future work will explore more sophisticated selection strategies to address this gap. Second, the two-stage training paradigm, while efficient in practice (first stage requires only $\sim$1 hour), may yield suboptimal coordination between stages compared to end-to-end alternatives.

\section{Conclusion}

In this work, we systematically analyze the challenges inherent in vast vocabulary detection: positive gradient dilution and hard negative gradient dilution. Through comprehensive experiments, we expose the limitations of Focal Loss under these challenging settings. To mitigate these issues, we propose CQ-DINO, a novel framework with two core innovations: (1) learnable category queries that encode category correlations, and (2) image-guided query selection that effectively reduces the negative space while performing adaptive hard negative mining. Extensive evaluations on the V3Det and COCO benchmarks demonstrate that CQ-DINO achieves superior performance and  strong scalability as vocabulary sizes increase. As future work, we plan to investigate the adaptability of our category query formulation for open-vocabulary detection and incremental learning scenarios.

\section*{Acknowledgments}
This work was supported in part by NSFC under Grant 62222112 and 62176186; and in part by the Innovative Research Group Project of Hubei Province under Grant
2024AFA017.

\newpage
{
\small
\bibliographystyle{plain}
\bibliography{main}

}

\appendix

\clearpage
\appendix
\section{Appendix}

\subsection{Extended Analysis on Gradient Dilution in Vast Vocabulary vs. Class-Imbalanced}
In Sec.~\ref{sec:challenge} of the main paper, we introduced the concept of positive gradient dilution as a primary challenge in vast vocabulary object detection, using a simplified model assuming a relatively balanced data distribution. This appendix provides a more comprehensive analysis to further clarify the distinction between the gradient dilution problem caused by vast vocabulary size and the one addressed in traditional class-imbalanced learning.

The simplified gradient ratio in our main paper (Eq.~\eqref{eq:gradient_ratio}) demonstrates that $\rho \propto \frac{1}{C}$, highlighting the direct impact of the vocabulary size $C$. To better distinguish the effects of class imbalance and vocabulary size, we can formulate a more general gradient signal ratio, $\rho_{c^+}$, for a positive class $c^+$:
\[
\rho_{c^+} = \frac{n_{c^+} \cdot \|\nabla_{z_{c^+}}\mathcal{L}\|}{\sum_{c^-\neq c^+}^C n_{c^-}\cdot\|\nabla_{z_{c^-}}\mathcal{L}\|} \approx \frac{n_{c^+}\cdot(1-\sigma(z_{c^+}))}{ \mathbb{E}_{c^-}[\sigma(z_{c^-})]\cdot\sum_{c^-\neq c^+}^C n_{c^-}}
\]
By letting $\epsilon^+$ and $\epsilon^-$ represent the average gradient magnitudes for positive and negative samples respectively, and $N$ be the total sample count across all classes, the ratio can be expressed as:

\begin{equation}
\rho_{c^+} \propto \frac{n_{c^+}\cdot \epsilon^+}{(N-n_{c^+})\cdot \epsilon^-}
\label{eq:appendix_grad_ratio_general}
\end{equation}
This generalized formula reveals two distinct and compounding sources of gradient dilution:

 \textbf{Class Imbalance:} This well-studied issue is primarily reflected by the term $n_{c^+}$ in the numerator. When a class is rare (i.e., has a long-tailed distribution), its sample count $n_{c^+}$ is small, which directly reduces the gradient signal ratio $\rho_{c^+}$. This is the central challenge that traditional class-imbalance methods aim to solve.

\textbf{Vast Vocabulary Size:} This is primarily driven by the term $N - n_{c^+}$ in the denominator. In vast vocabulary settings ($C>10,000$), $N - n_{c^+}$ becomes enormous because it aggregates all negative samples from the other $C-1$ categories. The large $C$ further reduces the ratio $\rho_{c^+}$, making the gradient dilution more severe in vast vocabulary object detection. This impacts \textbf{both rare and common classes}. 

This distinction in the problem's source explains why conventional methods for class imbalance are not sufficient for the vast vocabulary challenge.

\subsection{Algorithm Details}
This section provides the detailed algorithmic implementations of the three core components in CQ-DINO, as described in Sec.~\ref{method}. Each algorithm addresses specific challenges in vast vocabulary object detection:

\textbf{Algorithm 1 - Image-Guided Query Selection (Sec.~\ref{query_section}):} This algorithm implements the core innovation of CQ-DINO by dynamically selecting the most relevant category queries for each image. Through cross-attention mechanisms between category queries and image features, it reduces the negative search space from the full vocabulary to a manageable subset ($C' \ll C$), addressing both positive gradient dilution and hard negative gradient dilution issues identified in Sec.~\ref{sec:challenge}.

\begin{algorithm}[H]
\caption{Image-Guided Query Selection}
\label{alg:image_guided_query_selection}
\KwIn{ \\
    \quad\quad $B$: Batch size, $C$: Number of category queries, \\
    \quad\quad $D$: Embedding dimension, $C'$: Number of selected category queries, \\
    \quad\quad $H'$: Height of image features, $W'$: Width of image features,\\
    \quad\quad $\mathbf{Q}_{cat} \in \mathbb{R}^{B \times C \times D}$: Category queries,\\
    \quad\quad $\mathbf{F}_{img} \in \mathbb{R}^{B \times D \times H' \times W'}$: Image features%
}
\KwOut{\\
    \quad\quad $\mathbf{Q}_{cat}' \in \mathbb{R}^{B \times C' \times D}$: Selected enhanced category queries,\\
    \quad\quad $\mathcal{I} \in \mathbb{R}^{B \times C'}$: Selection indices%
}

\BlankLine
\hrule
\BlankLine

\tcpgreen{Reshape image features for cross-attention}
$\mathbf{F}_{flat} \leftarrow \text{Reshape}(\mathbf{F}_{img}, [B, D, H' \times W'])$\;
$\mathbf{Q}_{enhanced} \leftarrow \mathbf{Q}_{cat}$\;

\BlankLine
\tcpgreen{Enhancement through cross-attention layers}
\For{$l = 1$ \KwTo $2$}{
    \tcpgreen{Cross-attention with image features}
    $\mathbf{Q}_{attn} \leftarrow \text{MultiHeadCrossAttention}(\mathbf{Q}_{enhanced}, \mathbf{F}_{flat}, \mathbf{F}_{flat})$\;
    $\mathbf{Q}_{enhanced} \leftarrow \text{LayerNorm}(\mathbf{Q}_{attn} + \mathbf{Q}_{enhanced})$\;
    
    \tcpgreen{Feed-forward transformation}
    $\mathbf{Q}_{ffn} \leftarrow \text{FFN}(\mathbf{Q}_{enhanced})$\;
    $\mathbf{Q}_{enhanced} \leftarrow \text{LayerNorm}(\mathbf{Q}_{ffn} + \mathbf{Q}_{enhanced})$\;
}

\BlankLine
\tcpgreen{Query selection based on enhanced representations}
$\mathbf{L} \leftarrow \text{LinearProjection}(\mathbf{Q}_{enhanced})$ \tcpgreen{$\mathbf{L} \in \mathbb{R}^{B \times C}$}

\For{$b = 1$ \KwTo $B$}{
    $\mathcal{I}_{b,:} \leftarrow \text{TopK}(\mathbf{L}_{b,:}, C')$\;
    $\mathbf{Q}_{cat}'[b,:,:] \leftarrow \mathbf{Q}_{enhanced}[b, \mathcal{I}_{b,:}, :]$\;
}

\Return{$\mathbf{Q}_{cat}'$, $\mathcal{I}$}
\end{algorithm}

\textbf{Algorithm 2 - Self-Attention for Implicit Category Relations (Sec.~\ref{sec:tree_struct}):} For datasets without explicit hierarchical structures, this algorithm employs multi-head self-attention to learn implicit correlations between categories. It allows semantically related categories to influence each other's representations based on learned attention patterns, complementing the explicit hierarchical approach.

\begin{algorithm}[H]
\caption{Self-Attention for Implicit Category Relations}
\label{alg:self_attention_category_relations}
\KwIn{%
    $\mathbf{Q}_{cat} \in \mathbb{R}^{C \times D}$: Category queries,\\
    $H$: Number of attention heads%
}
\KwOut{$\mathbf{Q}_{corr} \in \mathbb{R}^{C \times D}$: Correlation-enhanced queries}

\BlankLine
\hrule
\BlankLine

\tcpgreen{Enhance queries through self-attention mechanism}
$\mathbf{Q}_{attn} \leftarrow \text{MultiHeadSelfAttention}(\mathbf{Q}_{cat})$\;
$\mathbf{Q}_{corr} \leftarrow \text{LayerNorm}(\mathbf{Q}_{attn} + \mathbf{Q}_{cat})$\;

\Return{$\mathbf{Q}_{corr}$}
\end{algorithm}

\textbf{Algorithm 3 - Explicit Hierarchical Tree Construction (Sec.~\ref{sec:tree_struct}):} For datasets with explicit hierarchical structures like V3Det, this algorithm leverages the category hierarchical structures to enhance query representations. It performs bottom-up tree traversal to incorporate hierarchical relationships through adaptive weighting, enabling parent categories to aggregate semantic information from their children while maintaining individual semantics.

\begin{algorithm}[H]
\caption{Explicit Hierarchical Tree Construction}
\label{alg:hierarchical_tree_construction}
\KwIn{%
    $\mathbf{Q}_{cat} \in \mathbb{R}^{C \times D}$: Category queries,\\
    $\mathcal{T}$: Hierarchical tree structure with nodes $\mathcal{V}$,\\
    $w$: Base weight parameter (default: 0.3)%
}
\KwOut{$\mathbf{Q}_{corr} \in \mathbb{R}^{C \times D}$: Correlation-enhanced queries}

\BlankLine
\hrule
\BlankLine

\tcpgreen{Initialize and prepare tree traversal}
$\mathbf{Q}_{corr} \leftarrow \mathbf{Q}_{cat}$\;
$N_{max} \leftarrow \max_{v \in \mathcal{V}} |\text{Children}(v, \mathcal{T})|$\;
$\mathcal{L} \leftarrow \text{TopologicalSort}(\mathcal{T})$\;

\BlankLine
\tcpgreen{Bottom-up tree traversal for correlation enhancement}
\ForEach{node $v$ in $\text{Reverse}(\mathcal{L})$}{
    \uIf{$\text{IsLeaf}(v, \mathcal{T})$}{
        \tcpgreen{Leaf nodes retain original queries}
        \textbf{continue}\;
    }
    \Else{
        \tcpgreen{Compute adaptive weight based on children count}
        $\mathcal{C}(v) \leftarrow \text{GetChildren}(v, \mathcal{T})$\;
        $n_v \leftarrow |\mathcal{C}(v)|$\;
        $\alpha_v \leftarrow w + \frac{\log(n_v + 1)}{\log(N_{max} + 1)}$\;
        $\alpha_v \leftarrow \min(\alpha_v, 1.0)$\;
        
        \BlankLine
        \tcpgreen{Update parent query with weighted combination}
        $\mathbf{Q}_{child}^{mean} \leftarrow \frac{1}{n_v} \sum_{c \in \mathcal{C}(v)} \mathbf{Q}_{corr}[c,:]$\;
        $\mathbf{Q}_{corr}[v,:] \leftarrow (1-\alpha_v) \cdot \mathbf{Q}_{cat}[v,:] + \alpha_v \cdot \mathbf{Q}_{child}^{mean}$\;
    }
}

\Return{$\mathbf{Q}_{corr}$}
\end{algorithm}

\subsection{Performance of DINO with Different Focal Loss Parameters}
In Tab.~\ref{tab:focal}, we achieve the optimal Focal Loss parameters at $\alpha = 0.5 $ and $\gamma = 2 $ with the Swin-B-22k backbone. We conduct parameter setting experiments with different backbone in Tab.~\ref{tab:dino}. We find that this optimal setting does not generalize well. The same parameters ($\alpha=0.5$, $\gamma=2$) lead to a performance degradation of 3.3\% AP for the Swin-B backbone compared to the default configuration ($\alpha=0.25$, $\gamma=2$). However, they improve the performance by 4.0\% AP and 1.6\% AP for the Swin-B-22k and Swin-L backbones, respectively. These findings show that while Focal loss addresses gradient dilution issues, its optimal configuration requires careful parameter tuning.

\begin{table}[h]
\centering
\small
\caption{Performance comparison between standard DINO ($\alpha=0.25$, $\gamma=2$), $\ddagger$~DINO with modified Focal loss parameters ($\alpha=0.50$, $\gamma=2$), and the proposed CQ-DINO.}
\label{tab:dino}
\begin{tabular}{@{}l lll l@{}}
\toprule

Method & Backbone & $AP$ & $AP_{50}$ & $AP_{75}$ \\
\midrule
DINO  & Swin-B & 42.0  & 46.8 & 43.9  \\
 $\ddagger$~DINO   & Swin-B & 38.7 \tiny{($\downarrow3.3$)}  & 43.7 \tiny{($\downarrow3.1$)} & 40.4 \tiny{($\downarrow3.5$)} \\
CQ-DINO  & Swin-B & \textbf{46.3}  & \textbf{51.5} & \textbf{48.4} \\
\midrule
DINO & Swin-B-22k & 43.4  & 48.4 & 45.4 \\
$\ddagger$~DINO & Swin-B-22k & 47.4 \tiny{($\uparrow4.0$)}   & 53.3 \tiny{($\uparrow4.9$)}  & 49.7 \tiny{($\uparrow4.3$)}  \\
CQ-DINO  & Swin-B-22k & \textbf{52.3}  & \textbf{57.7} & \textbf{54.6} \\
\midrule
DINO & Swin-L & 48.5 & 54.3 & 50.7 \\
$\ddagger$~DINO & Swin-L & 50.1 \tiny{($\uparrow1.6$)}  & 56.3 \tiny{($\uparrow2.0$)} & 52.4 \tiny{($\uparrow1.7$)} \\
CQ-DINO  & Swin-L & \textbf{53.0} & \textbf{58.4} & \textbf{55.4} \\

\bottomrule
\end{tabular}
\end{table}

\subsection{Gradient Norm Visualization During Training}
\begin{figure}[h]
    \centering
    \includegraphics[width=1.03\linewidth]{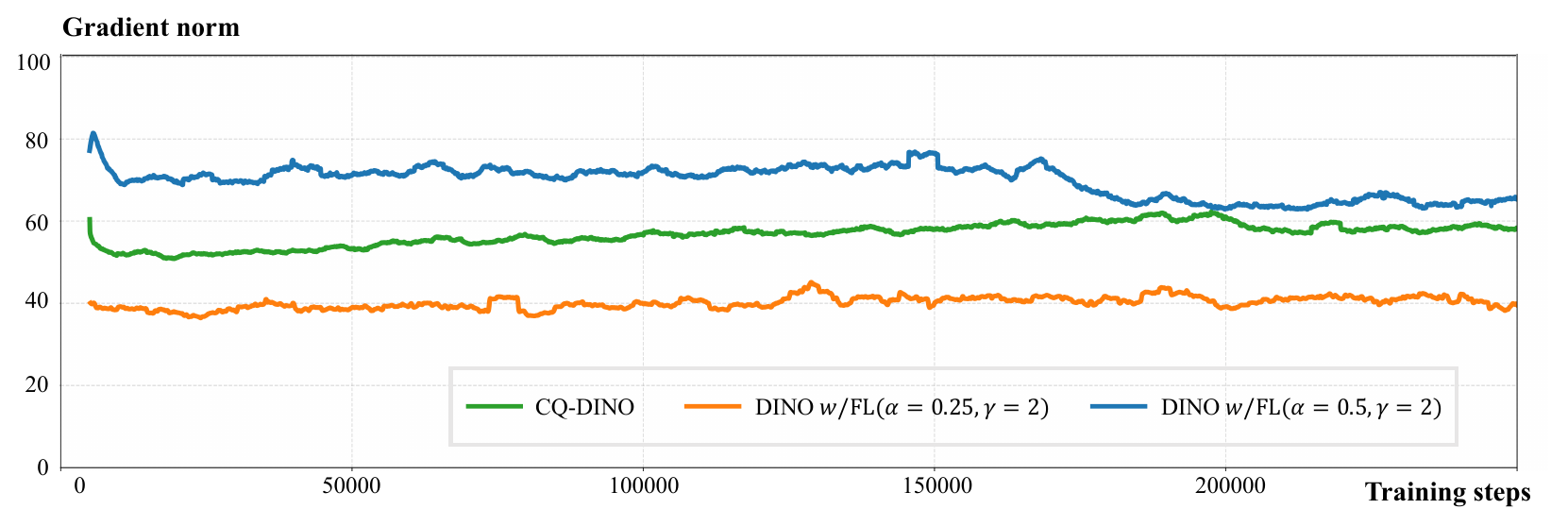}
\caption{Gradient norm visualization during training process.}
    \label{fig:gradient_vis}
\end{figure}

As reported in Tab.~\ref{tab:focal}, CQ-DINO achieves superior performance compared to DINO with optimal Focal Loss parameters ($\alpha=0.5$, $\gamma=2$), outperforming the default configuration of DINO ($\alpha=0.25$, $\gamma=2$). To analyze their impact on training gradients, we visualize the gradient norm in Fig.~\ref{fig:gradient_vis} for three configurations: (1) CQ-DINO (red), (2) DINO with $\alpha=0.5$ (blue), and (3) DINO with $\alpha=0.25$ (orange), all using the Swin-B-22k backbone. The results reveal that the gradient norm for DINO with $\alpha=0.25$ (orange line) remains low throughout training, indicating insufficient learning from both positive and hard negative samples. In contrast, DINO with $\alpha=0.5$ (blue line) initially displays strong gradients, but these are unstable and fluctuate considerably, as reflected in the high variance of the blue points. Meanwhile, CQ-DINO maintains a balanced trajectory, sustaining moderate and stable gradient magnitudes during the entire training process. This sustained and balanced gradient norm demonstrates that CQ-DINO more effectively addresses the gradient dilution issues.

\subsection{Extended Gradient Ratio Analysis}

\begin{figure}[h]
    \centering
    \includegraphics[width=1\linewidth]{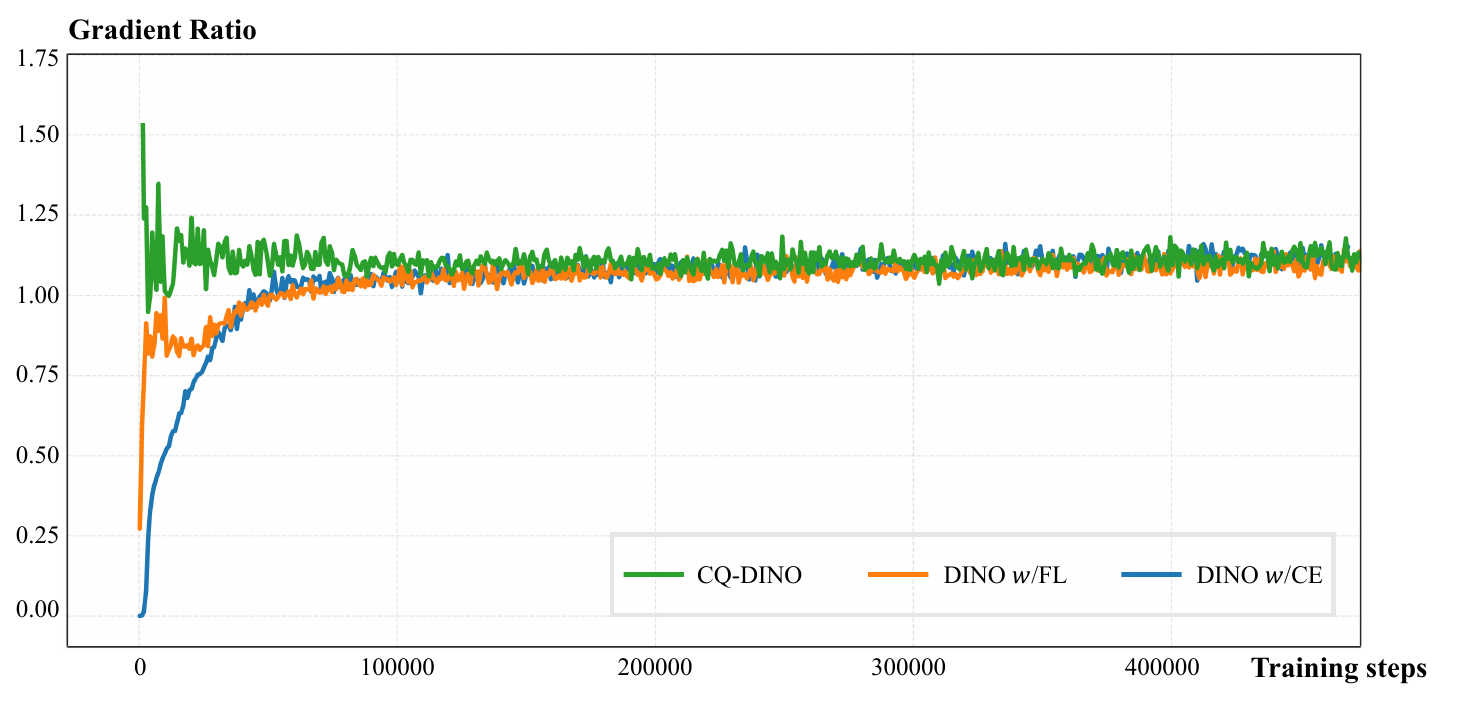}
\caption{Positive-to-negative gradient ratio across extended training iterations on V3Det dataset.}
    \label{fig:gradient_ratio}
\end{figure}

\paragraph{Positive-to-Negative Gradient Ratio.}
We extend our analysis to 400k training iterations to examine the effect of CQ-DINO on gradient distribution. As illustrated in Fig.~\ref{fig:gradient_ratio}, CQ-DINO mitigates the positive gradient dilution problem inherent in vast vocabulary tasks. During the early training stage (fewer than 10k iterations; see Fig.~\ref{fig:ratio}), CQ-DINO maintains a substantially higher positive-to-negative gradient ratio compared to DINO with cross-entropy (CE) or focal loss (FL). Moreover, CQ-DINO reaches a balanced state (>1.0) earlier than baselines such as DINO w/CE and DINO w/FL.

\begin{figure}[h]
    \centering
    \includegraphics[width=1\linewidth]{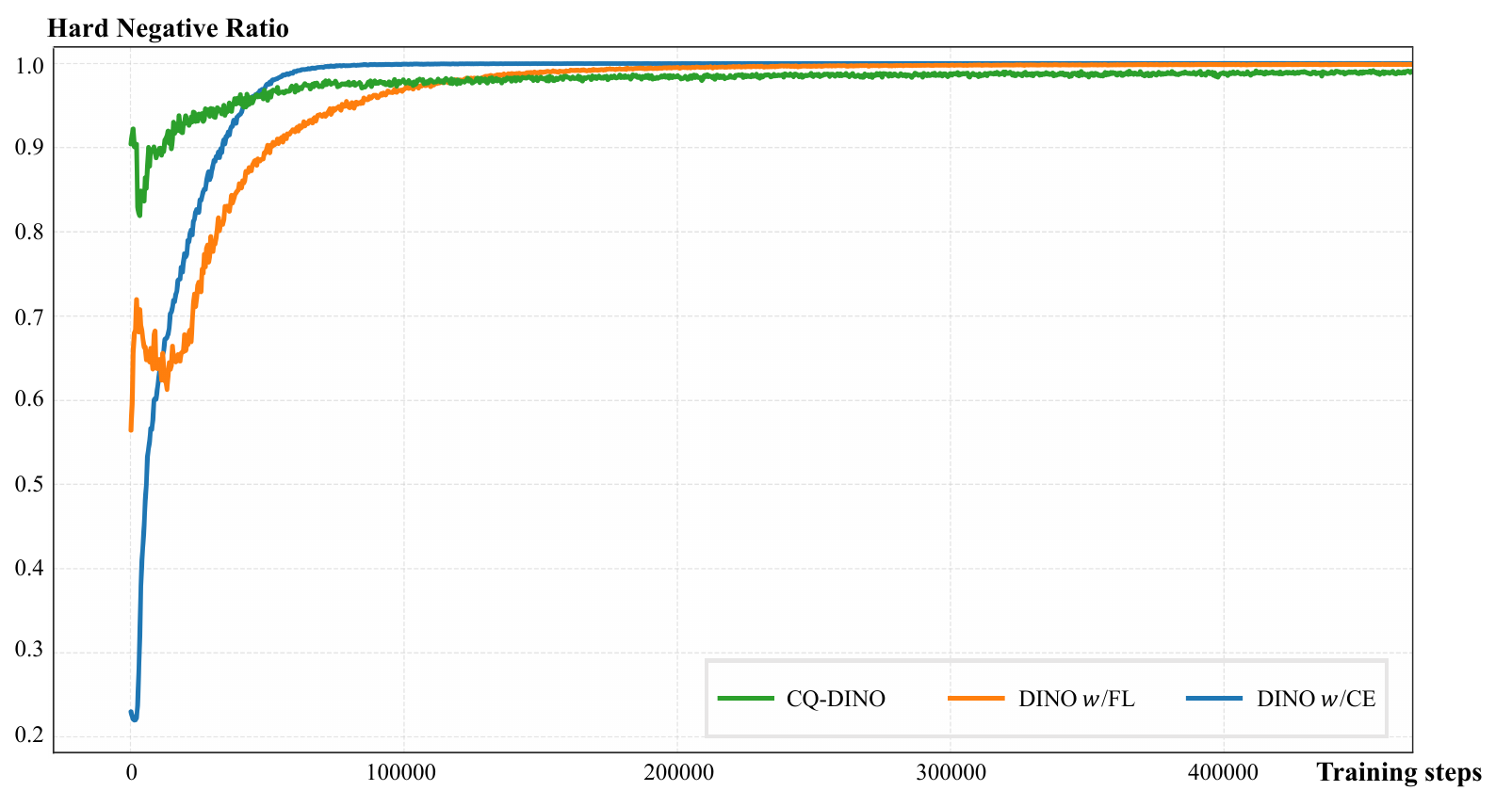}
\caption{Hard negative-to-all negative gradient ratio across training iterations on V3Det dataset.}
    \label{fig:hard_negative}
\end{figure}

\paragraph{Hard-Negative Gradient Contribution.}
We define hard negatives as the top 10\% of negative categories with the highest prediction scores, representing the most confusing negatives that require focused learning. Fig.~\ref{fig:hard_negative} presents the proportion of total negative-gradient magnitude attributable to hard negatives. In early training stage, CQ-DINO achieves higher hard-negative ratios compared to DINO w/CE and DINO w/FL. This confirms that our image-guided query selection implicitly performs effective hard-negative mining by filtering irrelevant categories and concentrating on informative ones.

\subsection{Additional Ablations on Isolated Component Contributions}

To provide a comprehensive understanding of our method's effectiveness and isolate the contributions of individual components, we conduct additional ablation studies that complement the analysis presented in Tab.~\ref{tab:component_ablation}. We analyze the following components, with results presented in Tab.~\ref{tab:r1}:

\begin{enumerate}
\item Baseline Architecture: The foundational model based on Grounding DINO, including its feature enhancer and decoder. This corresponds to the setting where all our proposed components are disabled (row 1).
\item Training-stage Gradient Dilution Mitigation: Our proposed image-guided query selection method applied during training. This ensures that the model's text encoder and fusion layers receive focused gradient signals from a small subset of relevant categories for each image.

\item Inference-stage Category Selection: A computational efficiency mechanism that selects only the top-K most relevant categories based on similarity between category queries and image features during inference.

\item  Encoding Category Correlations: the hierarchical tree construction method, which was identified as the effective approach for vast vocabulary setting.

\end{enumerate}

\begin{table}[h]
\centering
\caption{Detailed ablation study on the isolated contributions of our proposed components. ``0'' denotes the absence of the corresponding component, ``1'' denotes its inclusion.}
\label{tab:r1}
\small
\begin{tabular}{c|c|c|ccccc}
\toprule
   Training-stage & Inference-stage & Encoding category & 
    \multicolumn{2}{c}{V3Det} & \multicolumn{2}{c}{COCO} 
\\
   gradient dilution & category selection & correlations  &  AP &  FPS &  AP &  FPS \\

\midrule
0 & 0 & 0 & 47.3 & 0.7 & 57.5 & 9.8 \\
0 & 0 & 1 & 49.4 & 0.6 & 57.9 & 9.7 \\
0 & 1 & 1 & 46.6 & 10.6 & 58.2 & 9.9 \\
1 & 0 & 1 & 51.3 & 0.6 & 58.2 & 9.7 \\
1 & 1 & 0 & 51.1 & 10.8 & 58.3 & 10.0 \\
1 & 1 & 1 & \textbf{52.3} & 10.6 & \textbf{58.5} & 9.9 \\
\bottomrule
\end{tabular}
\end{table}

The detailed ablations in Tab.~\ref{tab:r1} confirm three main observations:

\textbf{Training-stage gradient dilution mitigation is the primary contributor.} On V3Det, introducing this component yields a $+1.9\%$ AP gain (row 2 vs.~row 4), validating our hypothesis that gradient dilution poses a significant challenge in large-vocabulary detection. On COCO, where the label space is much smaller, the improvement is modest ($+0.3\%$ AP).

\textbf{Inference-stage category selection alone does not resolve gradient dilution.} On V3Det, applying inference selection without training-stage mitigation reduces AP (row 2 vs.~row 3), likely because the underlying gradients remain diluted during training. In contrast, on COCO, inference selection provides a mild AP improvement due to reduced background competition.

\textbf{Combination leads to best accuracy–efficiency trade-off.} Integrating both training-stage gradient dilution mitigation and inference-stage category selection (row 6) achieves the highest AP on both datasets while providing substantial speedups: on V3Det, FPS improves from $0.6$ to $10.6$.

\subsection{Impact of Category Query Count}
Tab.~\ref{tab:number} presents the impact of varying the number of category queries (50, 100, 200) on performance. As expected, increasing the number of queries leads to higher $AR^C$ scores, at the cost of reduced inference speed. However, simply increasing the query count is not always beneficial. Excessive queries lead to gradient imbalance between positive and negative examples, resulting in diminishing performance gains. Furthermore, improvements in AR do not necessarily correlate with increases in  AP, as reflected in our observation that an $AR^C$ of 83.3\% is not the limiting factor at the current stage. Empirically, we find that 100 queries achieve an optimal balance between detection performance and computational efficiency for V3Det. On the COCO dataset, 30 queries are sufficient, achieving 99.1\% $AR^C$.\par

\begin{table}[h]
  \centering
  \caption{Ablation study with different numbers of selected category queries in CQ-DINO on V3Det.}
  \label{tab:number}
  \begin{tabular}{c | c c c @{}}
    \toprule
    Query Count & $AP$   &  $AR^{C}$ & FPS\\
    \midrule 
    50  & 51.8 & 78.4 & \textbf{6.9} \\
    100  & \textbf{52.3} & 83.3 & 6.6 \\
    200  & 52.2 & \textbf{87.5} & 5.9 \\
    \bottomrule
  \end{tabular}
\end{table}

\subsection{Failure Case Analysis}
To better understand the limitations of our approach and identify key bottlenecks, we conduct a systematic failure case analysis of CQ-DINO. Specifically, we examine the 3,527 categories for which the category-level recall falls below 83.3\%. This analysis highlights  category frequency and object scale as the dominant factors behind performance drops.

\paragraph{Analysis by Category Frequency.} Among the 3,527 low-recall categories, 374 are rare and 3,153 are common, following our defined frequency threshold. As shown in Tab.~\ref{tab:low_recall}, CQ-DINO achieves 20.5\% AP on rare categories, which is significantly lower than our overall performance of 52.3\% AP. This gap underscores the inherent difficulty of rare-category detection in large-vocabulary settings. Notably, despite the challenge, CQ-DINO consistently surpasses both the vanilla DINO and a DINO variant optimized with Focal Loss, indicating that the drop in rare-category performance is a domain-wide challenge rather than a limitation specific to our model design.

\begin{table}[ht]
\centering
\caption{Performance on category frequency with category-level recall below 83.3\%.}
\label{tab:low_recall}
\begin{tabular}{lccc}
\toprule
Method & Rare & Common & All \\
\midrule
DINO (Focal $\alpha$=0.50, $\gamma$=2) & 15.5 & 27.1 & 25.8 \\
DINO (Focal $\alpha$=0.50, $\gamma$=2) & 19.9 & 31.1 & 29.9 \\
\textbf{CQ-DINO} & \textbf{20.5} & \textbf{32.1} & \textbf{30.9} \\
\bottomrule
\end{tabular}
\end{table}

\paragraph{Analysis by Object Scale} We further assess performance across object scales (Tab.~\ref{tab:object_scale}). The largest gap appears for small objects, where CQ-DINO achieves only 14.1\% AP, far below the overall 52.3\% AP. While performance on medium- and large-scale objects is robust, this finding highlights that small object detection remains a critical challenge in vast vocabulary settings.

\begin{table}[ht]
\centering
\caption{Performance across object scales in low-recall cases.}
\label{tab:object_scale}
\begin{tabular}{lccc}
\toprule
Method & Small & Middle & Large \\
\midrule
DINO & 9.1 & 16.5 & 33.8 \\
DINO (Focal $\alpha$=0.50, $\gamma$=2) & 11.6 & 20.2 & 38.3 \\
\textbf{CQ-DINO} & \textbf{14.1} & \textbf{23.1} & \textbf{38.9} \\
\bottomrule
\end{tabular}
\end{table}


\newpage
\section*{NeurIPS Paper Checklist}

\begin{enumerate}

\item {\bf Claims}
    \item[] Question: Do the main claims made in the abstract and introduction accurately reflect the paper's contributions and scope?
    \item[] Answer: \answerYes{} 
    \item[] Justification: The abstract and introduction precisely reflect the scope and contributions of our work.
    \item[] Guidelines:
    \begin{itemize}
        \item The answer NA means that the abstract and introduction do not include the claims made in the paper.
        \item The abstract and/or introduction should clearly state the claims made, including the contributions made in the paper and important assumptions and limitations. A No or NA answer to this question will not be perceived well by the reviewers. 
        \item The claims made should match theoretical and experimental results, and reflect how much the results can be expected to generalize to other settings. 
        \item It is fine to include aspirational goals as motivation as long as it is clear that these goals are not attained by the paper. 
    \end{itemize}

\item {\bf Limitations}
    \item[] Question: Does the paper discuss the limitations of the work performed by the authors?
    \item[] Answer: \answerYes{} 
    \item[] Justification:  Section 4.6 provides a detailed discussion of our method's limitations.
    \item[] Guidelines:
    \begin{itemize}
        \item The answer NA means that the paper has no limitation while the answer No means that the paper has limitations, but those are not discussed in the paper. 
        \item The authors are encouraged to create a separate "Limitations" section in their paper.
        \item The paper should point out any strong assumptions and how robust the results are to violations of these assumptions (e.g., independence assumptions, noiseless settings, model well-specification, asymptotic approximations only holding locally). The authors should reflect on how these assumptions might be violated in practice and what the implications would be.
        \item The authors should reflect on the scope of the claims made, e.g., if the approach was only tested on a few datasets or with a few runs. In general, empirical results often depend on implicit assumptions, which should be articulated.
        \item The authors should reflect on the factors that influence the performance of the approach. For example, a facial recognition algorithm may perform poorly when image resolution is low or images are taken in low lighting. Or a speech-to-text system might not be used reliably to provide closed captions for online lectures because it fails to handle technical jargon.
        \item The authors should discuss the computational efficiency of the proposed algorithms and how they scale with dataset size.
        \item If applicable, the authors should discuss possible limitations of their approach to address problems of privacy and fairness.
        \item While the authors might fear that complete honesty about limitations might be used by reviewers as grounds for rejection, a worse outcome might be that reviewers discover limitations that aren't acknowledged in the paper. The authors should use their best judgment and recognize that individual actions in favor of transparency play an important role in developing norms that preserve the integrity of the community. Reviewers will be specifically instructed to not penalize honesty concerning limitations.
    \end{itemize}

\item {\bf Theory assumptions and proofs}
    \item[] Question: For each theoretical result, does the paper provide the full set of assumptions and a complete (and correct) proof?
    \item[] Answer: \answerYes{} 
    \item[] Justification: Complete proofs with all assumptions are provided in Sec. 3.1, and supporting details appear in Fig. 2, Fig. 6, Tab. 9, and Tab.10 ; all statements and proofs are clearly referenced.

    \item[] Guidelines:
    \begin{itemize}
        \item The answer NA means that the paper does not include theoretical results. 
        \item All the theorems, formulas, and proofs in the paper should be numbered and cross-referenced.
        \item All assumptions should be clearly stated or referenced in the statement of any theorems.
        \item The proofs can either appear in the main paper or the supplemental material, but if they appear in the supplemental material, the authors are encouraged to provide a short proof sketch to provide intuition. 
        \item Inversely, any informal proof provided in the core of the paper should be complemented by formal proofs provided in appendix or supplemental material.
        \item Theorems and Lemmas that the proof relies upon should be properly referenced. 
    \end{itemize}

    \item {\bf Experimental result reproducibility}
    \item[] Question: Does the paper fully disclose all the information needed to reproduce the main experimental results of the paper to the extent that it affects the main claims and/or conclusions of the paper (regardless of whether the code and data are provided or not)?
    \item[] Answer: \answerYes{} 
    \item[] Justification: We provide comprehensive information in Sec.4.2 to enable reproducibility of our main experimental results. Additionally, the code to reproduce our results are made available in the supplemental material. 
    \item[] Guidelines:
    \begin{itemize}
        \item The answer NA means that the paper does not include experiments.
        \item If the paper includes experiments, a No answer to this question will not be perceived well by the reviewers: Making the paper reproducible is important, regardless of whether the code and data are provided or not.
        \item If the contribution is a dataset and/or model, the authors should describe the steps taken to make their results reproducible or verifiable. 
        \item Depending on the contribution, reproducibility can be accomplished in various ways. For example, if the contribution is a novel architecture, describing the architecture fully might suffice, or if the contribution is a specific model and empirical evaluation, it may be necessary to either make it possible for others to replicate the model with the same dataset, or provide access to the model. In general. releasing code and data is often one good way to accomplish this, but reproducibility can also be provided via detailed instructions for how to replicate the results, access to a hosted model (e.g., in the case of a large language model), releasing of a model checkpoint, or other means that are appropriate to the research performed.
        \item While NeurIPS does not require releasing code, the conference does require all submissions to provide some reasonable avenue for reproducibility, which may depend on the nature of the contribution. For example
        \begin{enumerate}
            \item If the contribution is primarily a new algorithm, the paper should make it clear how to reproduce that algorithm.
            \item If the contribution is primarily a new model architecture, the paper should describe the architecture clearly and fully.
            \item If the contribution is a new model (e.g., a large language model), then there should either be a way to access this model for reproducing the results or a way to reproduce the model (e.g., with an open-source dataset or instructions for how to construct the dataset).
            \item We recognize that reproducibility may be tricky in some cases, in which case authors are welcome to describe the particular way they provide for reproducibility. In the case of closed-source models, it may be that access to the model is limited in some way (e.g., to registered users), but it should be possible for other researchers to have some path to reproducing or verifying the results.
        \end{enumerate}
    \end{itemize}

\item {\bf Open access to data and code}
    \item[] Question: Does the paper provide open access to the data and code, with sufficient instructions to faithfully reproduce the main experimental results, as described in supplemental material?
    \item[] Answer: \answerYes{} 
    \item[] Justification: The codes are fully available in the supplemental material.
    \item[] Guidelines:
    \begin{itemize}
        \item The answer NA means that paper does not include experiments requiring code.
        \item Please see the NeurIPS code and data submission guidelines (\url{https://nips.cc/public/guides/CodeSubmissionPolicy}) for more details.
        \item While we encourage the release of code and data, we understand that this might not be possible, so “No” is an acceptable answer. Papers cannot be rejected simply for not including code, unless this is central to the contribution (e.g., for a new open-source benchmark).
        \item The instructions should contain the exact command and environment needed to run to reproduce the results. See the NeurIPS code and data submission guidelines (\url{https://nips.cc/public/guides/CodeSubmissionPolicy}) for more details.
        \item The authors should provide instructions on data access and preparation, including how to access the raw data, preprocessed data, intermediate data, and generated data, etc.
        \item The authors should provide scripts to reproduce all experimental results for the new proposed method and baselines. If only a subset of experiments are reproducible, they should state which ones are omitted from the script and why.
        \item At submission time, to preserve anonymity, the authors should release anonymized versions (if applicable).
        \item Providing as much information as possible in supplemental material (appended to the paper) is recommended, but including URLs to data and code is permitted.
    \end{itemize}

\item {\bf Experimental setting/details}
    \item[] Question: Does the paper specify all the training and test details (e.g., data splits, hyperparameters, how they were chosen, type of optimizer, etc.) necessary to understand the results?
    \item[] Answer: \answerYes{} 
    \item[] Justification: Detailed implementation details are provided in the Sec. 4.2.
    \item[] Guidelines:
    \begin{itemize}
        \item The answer NA means that the paper does not include experiments.
        \item The experimental setting should be presented in the core of the paper to a level of detail that is necessary to appreciate the results and make sense of them.
        \item The full details can be provided either with the code, in appendix, or as supplemental material.
    \end{itemize}

\item {\bf Experiment statistical significance}
    \item[] Question: Does the paper report error bars suitably and correctly defined or other appropriate information about the statistical significance of the experiments?
    \item[] Answer: \answerNo{} 
    \item[] Justification: We ensured reproducibility with fixed seeds. Besides, we found that results varied minimally across multiple random seeds, though error bars or significance tests were not reported.
    \item[] Guidelines:
    \begin{itemize}
        \item The answer NA means that the paper does not include experiments.
        \item The authors should answer "Yes" if the results are accompanied by error bars, confidence intervals, or statistical significance tests, at least for the experiments that support the main claims of the paper.
        \item The factors of variability that the error bars are capturing should be clearly stated (for example, train/test split, initialization, random drawing of some parameter, or overall run with given experimental conditions).
        \item The method for calculating the error bars should be explained (closed form formula, call to a library function, bootstrap, etc.)
        \item The assumptions made should be given (e.g., Normally distributed errors).
        \item It should be clear whether the error bar is the standard deviation or the standard error of the mean.
        \item It is OK to report 1-sigma error bars, but one should state it. The authors should preferably report a 2-sigma error bar than state that they have a 96\% CI, if the hypothesis of Normality of errors is not verified.
        \item For asymmetric distributions, the authors should be careful not to show in tables or figures symmetric error bars that would yield results that are out of range (e.g. negative error rates).
        \item If error bars are reported in tables or plots, The authors should explain in the text how they were calculated and reference the corresponding figures or tables in the text.
    \end{itemize}

\item {\bf Experiments compute resources}
    \item[] Question: For each experiment, does the paper provide sufficient information on the computer resources (type of compute workers, memory, time of execution) needed to reproduce the experiments?
    \item[] Answer: \answerYes{} 
    \item[] Justification: We report inference times in Table 4, using 8 NVIDIA A100-40G GPUs for most experiments and 8 NVIDIA H800-80G GPUs for those in Table 4 due to memory constraints, providing sufficient detail to reproduce our compute setup.
    \item[] Guidelines:
    \begin{itemize}
        \item The answer NA means that the paper does not include experiments.
        \item The paper should indicate the type of compute workers CPU or GPU, internal cluster, or cloud provider, including relevant memory and storage.
        \item The paper should provide the amount of compute required for each of the individual experimental runs as well as estimate the total compute. 
        \item The paper should disclose whether the full research project required more compute than the experiments reported in the paper (e.g., preliminary or failed experiments that didn't make it into the paper). 
    \end{itemize}
    
\item {\bf Code of ethics}
    \item[] Question: Does the research conducted in the paper conform, in every respect, with the NeurIPS Code of Ethics \url{https://neurips.cc/public/EthicsGuidelines}?
    \item[] Answer: \answerYes{} 
    \item[] Justification: Our research fully complies with the NeurIPS Code of Ethics in all respects. No sensitive data or personally identifiable information was used. There are no conflicts of interest or ethical concerns in this work.
    \item[] Guidelines:
    \begin{itemize}
        \item The answer NA means that the authors have not reviewed the NeurIPS Code of Ethics.
        \item If the authors answer No, they should explain the special circumstances that require a deviation from the Code of Ethics.
        \item The authors should make sure to preserve anonymity (e.g., if there is a special consideration due to laws or regulations in their jurisdiction).
    \end{itemize}

\item {\bf Broader impacts}
    \item[] Question: Does the paper discuss both potential positive societal impacts and negative societal impacts of the work performed?
    \item[] Answer: \answerNA{} 
    \item[] Justification: The research presented in this paper remains purely at the experimental levels and has not yet results in any societal impacts.
    \item[] Guidelines:
    \begin{itemize}
        \item The answer NA means that there is no societal impact of the work performed.
        \item If the authors answer NA or No, they should explain why their work has no societal impact or why the paper does not address societal impact.
        \item Examples of negative societal impacts include potential malicious or unintended uses (e.g., disinformation, generating fake profiles, surveillance), fairness considerations (e.g., deployment of technologies that could make decisions that unfairly impact specific groups), privacy considerations, and security considerations.
        \item The conference expects that many papers will be foundational research and not tied to particular applications, let alone deployments. However, if there is a direct path to any negative applications, the authors should point it out. For example, it is legitimate to point out that an improvement in the quality of generative models could be used to generate deepfakes for disinformation. On the other hand, it is not needed to point out that a generic algorithm for optimizing neural networks could enable people to train models that generate Deepfakes faster.
        \item The authors should consider possible harms that could arise when the technology is being used as intended and functioning correctly, harms that could arise when the technology is being used as intended but gives incorrect results, and harms following from (intentional or unintentional) misuse of the technology.
        \item If there are negative societal impacts, the authors could also discuss possible mitigation strategies (e.g., gated release of models, providing defenses in addition to attacks, mechanisms for monitoring misuse, mechanisms to monitor how a system learns from feedback over time, improving the efficiency and accessibility of ML).
    \end{itemize}
    
\item {\bf Safeguards}
    \item[] Question: Does the paper describe safeguards that have been put in place for responsible release of data or models that have a high risk for misuse (e.g., pretrained language models, image generators, or scraped datasets)?
    \item[] Answer: \answerNA{} 
    \item[] Justification: The work does not involve data or models with high risk for misuse, so no additional safeguards are necessary.
    \item[] Guidelines:
    \begin{itemize}
        \item The answer NA means that the paper poses no such risks.
        \item Released models that have a high risk for misuse or dual-use should be released with necessary safeguards to allow for controlled use of the model, for example by requiring that users adhere to usage guidelines or restrictions to access the model or implementing safety filters. 
        \item Datasets that have been scraped from the Internet could pose safety risks. The authors should describe how they avoided releasing unsafe images.
        \item We recognize that providing effective safeguards is challenging, and many papers do not require this, but we encourage authors to take this into account and make a best faith effort.
    \end{itemize}

\item {\bf Licenses for existing assets}
    \item[] Question: Are the creators or original owners of assets (e.g., code, data, models), used in the paper, properly credited and are the license and terms of use explicitly mentioned and properly respected?
    \item[] Answer: \answerYes{} 
    \item[] Justification: All assets used in this work are publicly available for academic research. We have cited the original sources as required and, to the best of our knowledge, have respected their terms of use.
    \item[] Guidelines:
    \begin{itemize}
        \item The answer NA means that the paper does not use existing assets.
        \item The authors should cite the original paper that produced the code package or dataset.
        \item The authors should state which version of the asset is used and, if possible, include a URL.
        \item The name of the license (e.g., CC-BY 4.0) should be included for each asset.
        \item For scraped data from a particular source (e.g., website), the copyright and terms of service of that source should be provided.
        \item If assets are released, the license, copyright information, and terms of use in the package should be provided. For popular datasets, \url{paperswithcode.com/datasets} has curated licenses for some datasets. Their licensing guide can help determine the license of a dataset.
        \item For existing datasets that are re-packaged, both the original license and the license of the derived asset (if it has changed) should be provided.
        \item If this information is not available online, the authors are encouraged to reach out to the asset's creators.
    \end{itemize}

\item {\bf New assets}
    \item[] Question: Are new assets introduced in the paper well documented and is the documentation provided alongside the assets?
    \item[] Answer: \answerYes{} 
    \item[] Justification: We introduce a new model with comprehensive documentation provided in the supplemental material, including a structured README detailing data collection, preprocessing, usage, license, limitations, and anonymization, in compliance with double-blind review requirements.
    \item[] Guidelines:
    \begin{itemize}
        \item The answer NA means that the paper does not release new assets.
        \item Researchers should communicate the details of the dataset/code/model as part of their submissions via structured templates. This includes details about training, license, limitations, etc. 
        \item The paper should discuss whether and how consent was obtained from people whose asset is used.
        \item At submission time, remember to anonymize your assets (if applicable). You can either create an anonymized URL or include an anonymized zip file.
    \end{itemize}

\item {\bf Crowdsourcing and research with human subjects}
    \item[] Question: For crowdsourcing experiments and research with human subjects, does the paper include the full text of instructions given to participants and screenshots, if applicable, as well as details about compensation (if any)? 
    \item[] Answer: \answerNA{} 
    \item[] Justification: The paper does not involve crowdsourcing nor research with human subjects.
    \item[] Guidelines:
    \begin{itemize}
        \item The answer NA means that the paper does not involve crowdsourcing nor research with human subjects.
        \item Including this information in the supplemental material is fine, but if the main contribution of the paper involves human subjects, then as much detail as possible should be included in the main paper. 
        \item According to the NeurIPS Code of Ethics, workers involved in data collection, curation, or other labor should be paid at least the minimum wage in the country of the data collector. 
    \end{itemize}

\item {\bf Institutional review board (IRB) approvals or equivalent for research with human subjects}
    \item[] Question: Does the paper describe potential risks incurred by study participants, whether such risks were disclosed to the subjects, and whether Institutional Review Board (IRB) approvals (or an equivalent approval/review based on the requirements of your country or institution) were obtained?
    \item[] Answer: \answerNA{} 
    \item[] Justification: The paper dose not involve crowdsourcing nor research with human subjects.
    \item[] Guidelines:
    \begin{itemize}
        \item The answer NA means that the paper does not involve crowdsourcing nor research with human subjects.
        \item Depending on the country in which research is conducted, IRB approval (or equivalent) may be required for any human subjects research. If you obtained IRB approval, you should clearly state this in the paper. 
        \item We recognize that the procedures for this may vary significantly between institutions and locations, and we expect authors to adhere to the NeurIPS Code of Ethics and the guidelines for their institution. 
        \item For initial submissions, do not include any information that would break anonymity (if applicable), such as the institution conducting the review.
    \end{itemize}

\item {\bf Declaration of LLM usage}
    \item[] Question: Does the paper describe the usage of LLMs if it is an important, original, or non-standard component of the core methods in this research? Note that if the LLM is used only for writing, editing, or formatting purposes and does not impact the core methodology, scientific rigorousness, or originality of the research, declaration is not required.
    \item[] Answer: \answerNo{} 
    \item[] Justification: Large Language Models were used solely for writing and editing the paper, not as part of the research methods or experiments.
    \item[] Guidelines:
    \begin{itemize}
        \item The answer NA means that the core method development in this research does not involve LLMs as any important, original, or non-standard components.
        \item Please refer to our LLM policy (\url{https://neurips.cc/Conferences/2025/LLM}) for what should or should not be described.
    \end{itemize}

\end{enumerate}

\end{document}